%% file: main.tex
\documentclass[10pt,twocolumn,letterpaper]{article}

\usepackage{cvpr}              
\usepackage[accsupp]{axessibility}  

\input{preamble}
\definecolor{cvprblue}{rgb}{0.21,0.49,0.74}
\usepackage[pagebackref,breaklinks,colorlinks,allcolors=cvprblue]{hyperref}

\usepackage{bm}
\usepackage{bbm}

\def\paperID{39674} 
\def\confName{CVPR}
\def\confYear{2026}

\title{\underline{O}bject-\underline{C}entric \underline{V}ision \underline{T}oken \underline{P}runing for Vision Language Models}

\author{
Guangyuan Li\textsuperscript{1}\thanks{Equal contribution.}\quad
Rongzhen Zhao\textsuperscript{1}\thanks{Equal contribution; Corresponding author.}\quad
Jinhong Deng\textsuperscript{2}\quad
Yanbo Wang\textsuperscript{3}\quad
Joni Pajarinen\textsuperscript{1} \\
\textsuperscript{1}Aalto University \\
\textsuperscript{2}University of Electronic Science and Technology of China \\
\textsuperscript{3}Delft University of Technology \\
{\tt\small \{guangyuan.li, rongzhen.zhao, joni.pajarinen\}@aalto.fi}\\
{\tt\small jhdengvision@gmail.com}\quad
{\tt\small y.wang-27@tudelft.nl}
}

\usepackage{graphicx}      
\usepackage{amsmath,amssymb}
\usepackage{booktabs}      
\usepackage{multirow}

\begin{document}

\maketitle

\input{sec/0_abstract}

\input{sec/1_introduction}

\input{sec/2_related_work}

\input{sec/3_proposed_method}

\input{sec/4_experiment}

\input{sec/5_conclusion}

\section*{Acknowledgment}

We acknowledge the support of Finnish Center for Artificial Intelligence (FCAI), Research Council of Finland flagship program.
We also appreciate CSC - IT Center for Science, Finland, for granting access to supercomputers Mahti and Puhti, as well as LUMI, owned by the European High Performance Computing Joint Undertaking (EuroHPC JU) and hosted by CSC Finland in collaboration with the LUMI consortium.
Furthermore, we acknowledge the computational resources provided by the Aalto Science-IT project through the Triton cluster.

{
    \small
    \bibliographystyle{ieeenat_fullname}
    \bibliography{main}
}


\end{document}

%% file: sec/0_abstract.tex
\begin{abstract}
In Vision Language Models (VLMs), vision tokens are quantity-heavy yet information-dispersed compared with language tokens, thus consume too much unnecessary computation.
Pruning redundant vision tokens for high VLM inference efficiency has been continuously studied but all existing methods resort to indirect and non-guaranteed ways.
We propose OC-VTP, a direct and guaranteed approach to select the most representative vision tokens for high-efficiency yet accuracy-preserving VLM inference.
Our OC-VTP requires merely light-weight pre-training of a small object-centric vision token pruner, which can then be inserted into existing VLMs, without fine-tuning of any models on any datasets.
It is guaranteed that the most representative vision tokens are kept by minimizing the error in reconstructing the original unpruned tokens from the selected ones.
Across any vision pruning ratios, i.e., inference efficiency, our OC-VTP consistently helps mainstream VLMs to preserve the highest inference accuracy.
Our pruning also demonstrates interesting interpretability.
Our source code, model checkpoints and evaluation logs are available on https://github.com/GarryLarry010131/OC-VTP.
\end{abstract}

%% file: sec/1_introduction.tex
\section{Introduction}
\label{sec:introduction}

Recent Vision-Language Models (VLMs) have achieved strong performance on challenging multimodal tasks, particularly open-ended visual question understanding and answering \cite{llava, llavanext, qwen2.5-vl, mini-gemini, mini-gpt}. In typical vision-language inference, textual prompts are tokenized into dozens of language tokens, whereas an image is represented by hundreds or even thousands of vision tokens. Unlike language tokens, many vision tokens carry highly redundant information \cite{hiprune}, and often a small subset is sufficient to represent the whole. This motivates Vision Token Pruning (VTP) to reduce VLM computation burden and accelerate inference, while incurring little accuracy loss.


Given a computation budget, i.e., the number of tokens to be kept, \textbf{how to determine} which vision tokens are redundant?
ToMe \cite{ToMe} adopts the human intuition that \textit{similar} vision tokens are redundant.
FastV \cite{FastV} defines the importance as the average \textit{attention} score a vision token receives from all other vision tokens.
TRIM \cite{TRIM} uses the \textit{similarity} between vision tokens and the pooling of text tokens as the significance.
VisionZip \cite{VisionZip} also defines the dominance as the average \textit{attention} score or the attention score with the \texttt{\small{CLS}} token.  
SparseVLM \cite{SparseVLM} measure the significance as the \textit{attention} between vision tokens and relevant text tokens.
PyramidDrop \cite{PDrop} defines the importance as the \textit{attention} between vision tokens and the last text token.
HiPrune \cite{hiprune} uses the overall \textit{attention} scores of different vision layers to measure the importance of vision tokens. 

However, all these methods rely on \textbf{indirect criteria}.
Namely, various attention scores and similarities are meticulously designed as metrics, with the authors' ingenious \textit{intuition}, even based on their insightful \textit{observation}.
These are somehow effective, but without any guarantee that the most representative vision tokens are selects from all.

\input{images_tex/figure_teaser}


We formulate VTP \textbf{as an optimization problem} given the budget.
The \textit{ideal case} is that we prune the vision tokens that have the least effect on the vision-language inference accuracy. But this requires access to the text tokens, not easy to formalize for optimality guarantee analysis.
We focus on the \textit{practical case} that how to prune the least representative vision tokens that minimize the information loss from the original unpruned ones. This is straightforward to formalize for optimality guarantee analysis.

We propose Object-Centric Vision Token Pruning (OC-VTP).
It is the \textbf{first guaranteed method} that prunes the \textit{least representative} vision tokens, for high-efficiency yet accuracy-preserving VLM inference.
This is realized by pre-training a lightweight Object-Centric pruner (OC-pruner) on the vision tokens of a small subset of images to keep the most representative vision tokens that lose the least information. Then it can plug and play in various VLM inference.
As shown in \Cref{fig:teaser}, our OC-VTP outperforms state-of-the-art (SotA) VTP methods on various datasets in average accuracy given any budget.
Note that all our superiority is achieved without accessing the text tokens for aiding  the vision token pruning.

In summary, our contributions are: (\textit{i}) We realize guaranteed VTP for VLMs for the first time that can select the most representative tokens; (\textit{ii}) Our method requires lightweight pretraining of a lightweight module, which can be plugged into various VLMs without fine-tuning; (\textit{iii}) Our method achieves roughly new SotA in terms of VTP for VLM, with intuitive object-level interpretability.


%% file: images_tex/figure_teaser.tex


\begin{figure}
\centering
\includegraphics[width=0.6\linewidth]{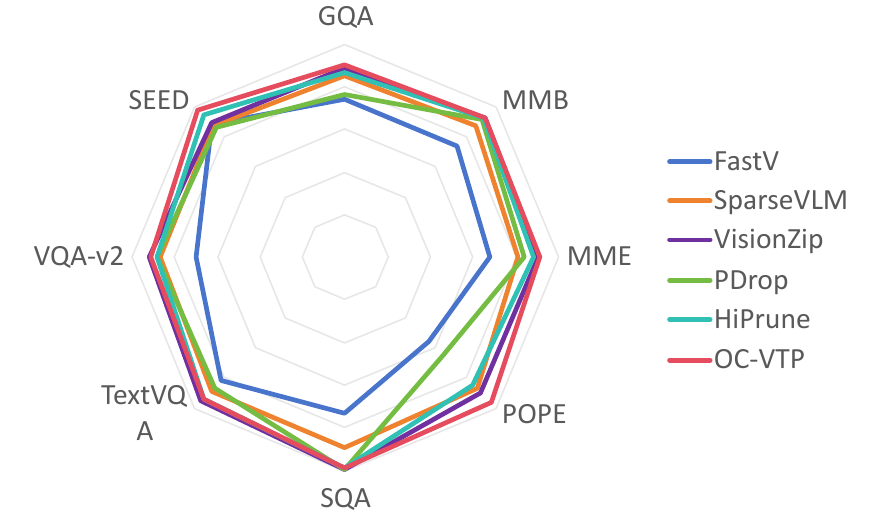}
\raisebox{1.5\baselineskip}{\includegraphics[width=0.39\linewidth]{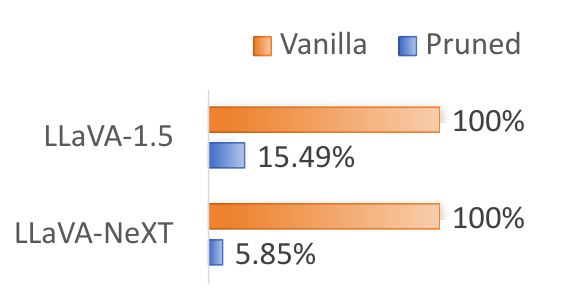}}
\caption{
(\textit{left}) Our OC-VTP consistently outperforms prior SotA methods, retaining over \textbf{95\%} of accuracy with only \textbf{11.1\%} of visual tokens on LLaVA-1.5.
(\textit{right}) Our OC-VTP reduces FLOPs by nearly \textbf{85\%} on LLaVA-1.5-7B when retaining \textbf{11.1\%} vision tokens, and by \textbf{95\%} on LLaVA-NeXT-7B when retaining \textbf{5.6\%} vision tokens, at a text length of 32, assuming MAC=2.
}
\label{fig:teaser}
\end{figure}




%% file: sec/2_related_work.tex
\section{Related Work}
\label{sec:related_work}

\subsection{Vision Language Models}

Mainstream VLMs consist of a vision encoder, a text tokenizer, a multimodal token projector, and a Large Language model (LLM) that infers upon such tokens.
LLaVA-1.5 \cite{llava} uses the CLIP \cite{CLIP} vision encoder for image tokenization, which produces 576 vision tokens. LLaVA-NeXT \cite{llavanext} adopts the same architecture, but targets high-resolution and multi-image inputs. Its AnyRes technique splits a high-resolution image into several sub-images, all are tokenized and then concatenated, producing 2880 ($5\times576$) tokens in total. Qwen2.5-VL \cite{Qwen2.0,qwen2.5-vl,ROPE} trains a native dynamic-resolution Vision Transformer (ViT) that keeps images at their original size while controlling cost.

These VLMs achieve strong results on benchmarks such as GQA \cite{gqa}, MMBench \cite{mmb}, POPE \cite{pope}, TextVQA \cite{textVQA} and ScienceQA \cite{sqa}. However, the inference cost is dominated by the number of vision tokens. High-resolution tiling and multi-image inputs further enlarge the inference computation and latency. This motivates pruning vision tokens, as text tokens are more information intensive, to reduce computation burden while preserving accuracy.

\subsection{Vision Token Pruning}
\label{sect:vision_token_pruning}
ToMe~\cite{ToMe} merges similar vision tokens based on token similarities, which happens inside the vision encoder.
TRIM \cite{TRIM} uses the similarity between vision tokens and text tokens' pooling to select tokens, while merging the remaining. Its VTP happens after the vision encoder and before the LLM decoder.
VisionZip \cite{VisionZip} uses the average or the attention with the \texttt{\small{CLS}} token to choose tokens, while group merging the remaining. Its VTP applies after the vision encoder and before the LLM decoder.
FastV \cite{FastV} uses the average attention score of a LLM decoder layer to identify important tokens and removes the remaining. Its VTP happens inside the LLM decoder. 
SparseVLM \cite{SparseVLM} relies on the attention between vision tokens and relevant text tokens to choose important tokens, while group merging the pruned tokens. Its VTP works inside the LLM decoder layers.
PyramidDrop \cite{PDrop} keeps tokens according to the attention between vision tokens and the last text token, while removing the remaining. Its VTP happens inside the LLM decoder layers.
HiPrune \cite{hiprune} utilizes the overall attention of different vision layers for pruning, while removing the remaining, after the vision encoder and before the LLM decoder.

Instead of relying on these handcrafted importance signal, our method learns a general optimality-guaranteed vision token pruner. Its VTP happens between the vision encoder and LLM decoder, no finetuning on VLMs needed.

\input{images_tex/method_overview}



\subsection{Object-Centric Learning}
\label{sect:object_centric_learning}

Object-Centric Learning (OCL) \cite{slotAttention} represents a scene as a few object-level feature vectors, i.e., \textit{slots}. OCL models mostly adopt the encoding-aggregation-decoding architecture and are trained in self-supervision that minimizes the error in reconstructing the original input from these a few slots. This aligns with the VTP objective.

For aggregation, Slot Attention \cite{slotAttention} is the core module of mainstream OCL, either on images like DINOSAUR \cite{DINOSAUR}, SlotDiffusion \cite{wu2023slotdiffuz}, SPOT \cite{kakogeorgiou2024spot} and VVO \cite{zhao2025vvo}, or on videos like VideoSAUR \cite{zadaianchuk2024videosaur}, SlotContrast \cite{manasyan2025slotcontrast} and RandSF.Q \cite{zhao2025randsfq}. 
Slot Attention and its variants, like BO-QSA \cite{jia2023boqsa}, which predefines the number of slots, ISA \cite{biza2023isa}, which learns geometric invariance, and SmoothSA \cite{zhao2025smoothsa}, which introduces a preheater, all ensure \textit{exclusiveness} and \textit{completeness} on the slots, i.e., least information redundancy or loss.
We choose the original version to build our OC-pruner, as it supports variant number of slots, does not change the feature distribution and is highly efficient.

For decoding, there are mixture decoders, like broadcast CNN \cite{slotAttention} and MLP \cite{DINOSAUR}, auto-regressive decoders, like the transformer \cite{kakogeorgiou2024spot} and random auto-regressive transformer \cite{zhao2025dias}, and denoising decoders, like the conditional Diffusion \cite{wu2023slotdiffuz}.
We choose the random auto-regressive transformer decoder to build our OC-pruner, as it is adopted by OCL SotA and also is the most lightweight solution.

%% file: images_tex/method_overview.tex
\begin{figure*}[t]
\centering
\includegraphics[width=0.98\linewidth]{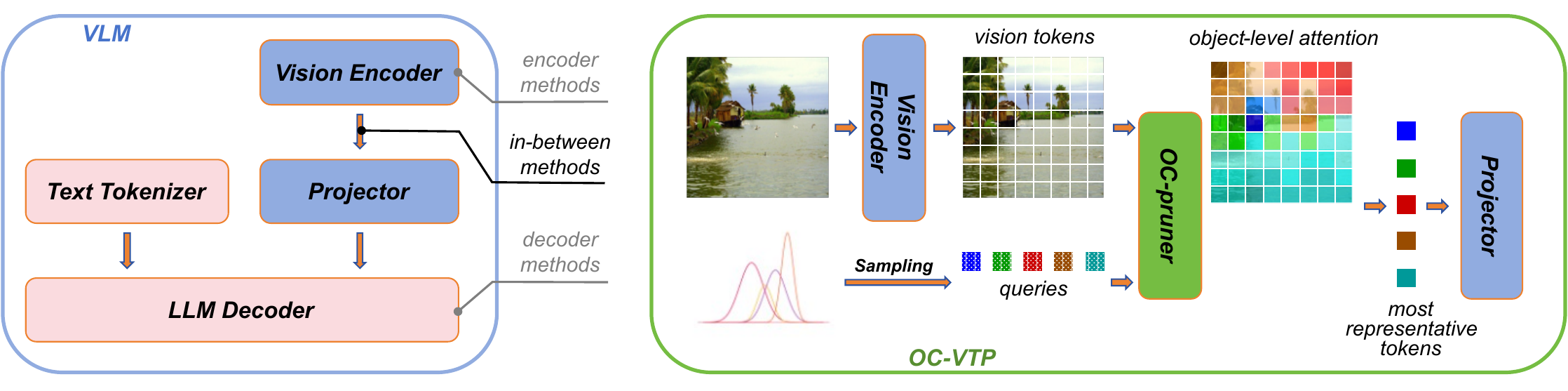}
\caption{
(\textit{left}) Structure of typical Vision-Language Models (VLMs), and three Vision Token Pruning (VTP) places:
\textit{encoder} methods like ToMe\cite{ToMe}; \textit{decoder} methods like SparseVLM, PyramidDrop, and FastV \cite{SparseVLM,PDrop, FastV}; and \textit{in-between} methods like TRIM, VisionZip and HiPrune \cite{TRIM,VisionZip,hiprune}.
Our OC-VTP operates in-between.
(\textit{right}) OC-pruner is the core of our OC-VTP.
It takes the middle layer tokens from the vision encoder as input, and employs queries sampled from a learned Gaussian distribution to locate the most representative token of each object or object part. The corresponding indexes are used to prune the vision tokens.
}
\label{fig:Method_Overview}
\end{figure*}

%% file: sec/3_proposed_method.tex
\section{Proposed Method}
\label{sect:proposed_method}

\subsection{Problem Formulation}
\label{sect:problem_formulation}

A vision-language inference task via a VLM has the form:
\input{equations/vl_task}

Here $\bm{\phi_\mathrm{VLM}}$ is the inference model. $\bm{V}$ and $\bm{L}$ are the vision and language parts of the inference task sample, respectively, with $\bm{L}$ mainly defining the task while $\bm{V}$ mainly providing the required information. $\bm{O}$ is the inference output and let us denote $\bm{Y}$ as the ground-truth answer.

Basically, the \textbf{ideal objective} of VTP for VLMs is to remove the vision tokens that have least impact on the inference task given the budget.
This requires us to develop an pruner that can prune vision tokens that cause the least average accuracy drop in all inference tasks:
\input{equations/ideal_objective}Here $\bm{\phi}_\mathrm{p}^*$ is the ideal pruner, keeping the most \textit{important} vision tokens $\bm{V}_\mathrm{p}^*$, according to the vision input $\bm{V}$ and language input $\bm{L}$, as well as the inference model $\bm{\phi}_\mathrm{VLM}$.

However, finding a solution for the ideal pruner $\bm{\phi}_\mathrm{p}^*$ submit to the ideal objective \Cref{eq:ideal_objective} is hardly possible. \textit{This is also why all existing VTP methods rely on handcrafted vision token importance metrics}.

But we can simplify this optimization problem into a \textbf{practical objective}. Assuming no access to the language input and the VLM, we need to develop a practical pruner that can keep vision tokens that preserve the most information of the original unpruned vision tokens:
\input{equations/practical_objective}Here $\bm{\phi}_\mathrm{p}$ is the practical pruner, which identifies the most \textit{representative} vision tokens $\bm{V}_\mathrm{p}$ from the original unpruned ones $\bm{V}$. $\bm{\phi}_\mathrm{d}$ is some distance metric, which measures the information gap between two token sets.

In this case, the practical pruner $\bm{\phi}_\mathrm{p}$ is agnostic to both language input $\bm{L}$ and the inference model $\bm{\phi}_\mathrm{VLM}$.
\textit{Finding a solution for it is quite feasible} as we only need to deal with the vision input $\bm{V}$ itself.


\subsection{Building OC-Pruner}



Based on \Cref{eq:practical_pruner}, we \textbf{design} our OC-pruner as follows. 
Given the budget, i.e., the vision sequence length after pruning, we draw the corresponding number of queries to aggregate the vision tokens into slots through a Slot Attention module, where vision tokens belonging to the same object or object part are aggregated into one slot.
\input{equations/oc_pruner_sa}Here $\bm{Q} \in \mathbb{R} ^ {s \times c}$ is the queries sampled from a learnt Gaussian distribution, with $s$ as the budget.
$\bm{V} \in \mathbb{R} ^ {n \times c}$ is the original vision tokens to be pruned.
$\bm{\phi}_\mathrm{SA}$ is the trained Slot Attention \cite{slotAttention} module, and please refer to \Cref{sect:object_centric_learning} for why not choosing other variants.
$\bm{S} \in \mathbb{R} ^ {s \times c}$ is the aggregated slots, with $\bm{A} \in \mathbb{R} ^ {s \times n}$ as the corresponding attention maps.
All these slots together can represent the original unpruned vision tokens with least information loss.

By the attention map of each slot, we can locate the most attended vision token, which is also the most representative token, then we can conduct the pruning:
\input{equations/oc_pruner_amax}\input{equations/oc_pruner_idx}Here the most attended indexes $\bm{I} \in \mathbb{R} ^ s$ are obtained by $\mathrm{argmax}$ $\bm{A}$ along the dimension $n$.
Besides $\mathrm{argmax}$, we also tried top-$k$, which as shown in \Cref{tab:ablation_topk} is inferior.
The pruned vision tokens $\bm{V}_\mathrm{p} \in \mathbb{R} ^ {s \times c}$ are obtained by selecting tokens from the original unpruned ones $\bm{V}$ with indexes $\bm{I}$.
Briefly, we implement the practical pruner $\bm{\phi}_\mathrm{p}$ as $\bm{\phi}_\mathrm{SA} \circ \mathrm{argmax} \circ [\cdot, :]$.

\subsection{Training OC-pruner}

Based on \Cref{eq:practical_objective}, we \textbf{train} our OC-pruner, especially the core module, Slot Attention, by firstly reconstructing the unpruned vision tokens from the slots and then by minimizing the reconstruction error:
\input{equations/oc_pruner_train}Here $\bm{V}' \in \mathbb{R} ^ {n \times c}$ is the reconstruction.
$\bm{\phi}_\mathrm{RAR}$ is the random auto-regressive transformer decoder \cite{zhao2025dias}.
As for why we do not choose other decoders, please refer to \Cref{sect:object_centric_learning} for detailed reasons.
$\bm{\phi}_\mathrm{r}$ is the reconstruction loss.
Briefly, we implement the distance metric $\bm{\phi}_\mathrm{d}$ as $\bm{\phi}_\mathrm{RAR} \circ \bm{\phi}_\mathrm{r}$.

\textbf{How to support any budget by training once}?
Mainstream VTP methods for VLMs support various pruning rates or budgets naturally, as they do not rely on learnt pruning criteria like our method.
To overcome such an issue, we train OC-pruner with \#slots queries $s$ randomly sampled from typical budget values, e.g., $\{32, 64, 128, 192\}$.

\textbf{How to preserve small yet important contents}?
Such reconstruction-based training objective \Cref{eq:oc_pruner_mse} minimizes the overall loss, thus vision tokens that are spatially small but contain important information might be ignored in the aggregation and decoding.
To address it, we propose our novel Area-Weighted Mean-Squared Error (AW-MSE):
\input{equations/AWMSE}Here segmentation masks $\bm{M} \in \mathbb{R} ^ {s \times n}$ corresponding to slots $\bm{S}$ are calculated by $\max$ along the dimension $s$. And the area of each mask, i.e., $\sum_{n_j} \bm{M}_{[s_i, n_j]}$, is used to reciprocally weight the corresponding squared error, i.e, $|| \bm{V}' - \bm{V} || ^ 2 _ {[\bm{M}_{[s_i, n_i]}, c_i]}$. Please refer to \Cref{tab:ablation_loss} for the effects of MSE vs AW-MSE on the performance.

\textbf{How to generalize to any datasets by training once}? 
Object-centric representation is good at zero-shot generalization \cite{DINOSAUR}. So we naively pre-train our OC-pruner on \textbf{40,000} images randomly sampled from COCO \cite{COCO}, i.e., no direct overlap with the evaluation.
For training efficiency, we preprocess these images into vision tokens as the training data via VLM vision encoders.
Despite evaluation images could follow quite different distributions, like synthetic vs real-world, even optical characters, our method, as shown in \Cref{sect:experiment}, performs surprisingly steadily.

\textbf{Comment}. 
The competition mechanism inside Slot Attention \cite{slotAttention} intrinsically ensures the kept vision tokens have least information overlap, i.e., \textit{exclusiveness}, while our training objective \Cref{eq:oc_pruner_mse} ensures the union of the kept tokens preserve the most information from the original unpruned tokens, i.e., \textit{completeness}.
These two aspects guarantee the most representative vision tokens are kept.
In contrast, as discussed in \Cref{sect:vision_token_pruning}, all existing methods rely on their handcrafted attention or similarity, which are not designed and optimized for such guarantee.

\subsection{Instantiating OC-VTP}

As shown in \Cref{fig:Method_Overview}, we insert the learnt OC-Pruner between the vision encoder and projector.
Beside, the vision tokens from some middle layer of the vision encoder should also be fed into OC-pruner as the pruning reference.
This is because to prune the vision tokens (from the layer), taking the middle layer vision tokens as the reference is better than taking themselves as the reference.
Namely, the input $\bm{V}$ to the Slot Attention module $\bm{\phi}_\mathrm{SA}$ in \Cref{eq:oc_pruner_sa} should be replace with some middle layer vision tokens, while the input $\bm{V}$ to the selection operation in \Cref{eq:oc_pruner_idx} should still be the last layer vision tokens.
Please refer to \Cref{tab:ablation_layer} for the effects of difference middle layers as reference.

\textbf{Comment}. Our OC-VTP operates between the vision encoder and projector.
In contrast, as discussed in \Cref{sect:vision_token_pruning}, some SotA methods \cite{ToMe} take first-mover advantage by pruning vision tokens inside the vision encoder to save computation earlier and more; some SotA methods \cite{SparseVLM,PDrop,FastV} have access to the language tokens inside the LLM decoder for more task-aware, rather than task-agonistic pruning.
Although theoretically possible, our OC-VTP, working without such techniques, still demonstrate consistent superiority, as shown in \Cref{sect:experiment}.

%% file: equations/vl_task.tex
\begin{equation}
\label{eq:vl_task}
\small
\bm{\phi}_\mathrm{VLM} : ( \bm{V} , \bm{L} ) \rightarrow \bm{O}
\end{equation}

%% file: equations/ideal_objective.tex
\begin{equation}
\small
\label{eq:ideal_objective}
\min 
\mathbb{E} _ {(\bm{V}, \bm{L})} \big[ \mathbbm{1} [ \bm{O} = \bm{Y} ] \big]
-
\mathbb{E} _ {(\bm{V}_\mathrm{p}^*, \bm{L})} \big[ \mathbbm{1} [ \bm{O}_\mathrm{p}^* = \bm{Y} ] \big]
\\
\end{equation}
\begin{equation}
\small
\label{eq:ideal_pruner}
\text{where\quad} \bm{\phi} _\mathrm{p} ^ * : ( \bm{V} ; \bm{L} , \bm{\phi}_\mathrm{VLM} ) \rightarrow \bm{V}_\mathrm{p}^*
\end{equation}

%% file: equations/practical_objective.tex
\begin{equation}
\small
\label{eq:practical_objective}
\min \bm{\phi}_\mathrm{d} ( \bm{V} , \bm{V}_\mathrm{p} )
\end{equation}
\begin{equation}
\small
\label{eq:practical_pruner}
\text{where\quad} \bm{\phi}_\mathrm{p} : \bm{V} \rightarrow \bm{V}_\mathrm{p}
\end{equation}

%% file: equations/oc_pruner_sa.tex
\begin{equation}
\small
\label{eq:oc_pruner_sa}
\bm{S}, \bm{A} = \bm{\phi}_\mathrm{SA} ( \bm{Q}, \bm{V} )
\end{equation}

%% file: equations/oc_pruner_amax.tex
\begin{equation}
\small
\label{eq:oc_pruner_amax}
\bm{I} = \mathrm{argmax} _ n ( \bm{A} )
\end{equation}

%% file: equations/oc_pruner_idx.tex
\begin{equation}
\small
\label{eq:oc_pruner_idx}
\bm{V}_\mathrm{p} = \bm{V} _ { [\bm{I}, :] }
\end{equation}

%% file: equations/oc_pruner_train.tex
\begin{equation}
\small
\label{eq:oc_pruner_dec}
\bm{V}' = \bm{\phi}_\mathrm{RAR} ( \bm{V}_\mathrm{p} )
\end{equation}
\begin{equation}
\small
\label{eq:oc_pruner_mse}
\min \bm{\phi}_\mathrm{r} ( \bm{V}', \bm{V} )
\end{equation}

%% file: equations/awmse.tex
\begin{equation}
\label{eq:awmse}
\small
\bm{\phi}_{\mathrm{r}} ( \bm{V}', \bm{V} ) = \frac{1}{s n c} \sum _ {s_i, n_i, c_i} \frac{1}{ \sum\limits _ {n_j} \bm{M} _ {[s_i, n_j]} } || \bm{V}' - \bm{V} || ^ 2 _ {[\bm{M}_{[s_i, n_i]}, c_i]}
\end{equation}
\begin{equation}
\small
\label{eq:mask}
\text{where\quad} \bm{M} = \mathbbm{1} [\bm{A} = \mathrm{max} _ s(\bm{A}) ]
\end{equation}

%% file: sec/4_experiment.tex
\input{tables/llava-1.5-acc}
\section{Experiments}
\label{sect:experiment}
We applied the pre-trained OC-Pruner to LLaVA-1.5, LLaVA-Next, and Qwen2.5-VL to test the results \cite{llava,llavanext,qwen2.5-vl}. We use LMMs-Eval to run ten official benchmarks \cite{lmms}, including GQA, MMB, MME, POPE, SQA, VQA$^{\text{Text}}$, VQA$^{\text{v2}}$, VizWiz, MMMU, and SEED \cite{gqa, mmb, mme, pope, sqa, textVQA, vqav2, vizwiz, mmmu, seedBench}. We compare our method with existing SotA approaches, FastV, SparseVLM, VisionZip, PyramidDrop, and HiPrune \cite{FastV,SparseVLM,VisionZip,PDrop,hiprune}.

\subsection{Results on Image Understanding}
\subsubsection{Results on LLaVA-1.5.}
As shown in \Cref{tab:llava-1.5-acc}, our OC-Pruner on LLaVA-1.5 outperforms SotA methods on most benchmarks and achieves the highest average relative accuracy to the vanilla model. With only 11.1\% vision tokens, OC-VTP reaches 95.5\% of the vanilla accuracy, exceeding the second-best method VisionZip at 93.1\%. At 22.2\% tokens, the gap narrows but remains the best at 97.7\% compared to the second-best HiPrune at 97.4\%.
\textbf{Given the fact that our method is only trained on COCO partial images, which has no direct overlap with these benchmarks, it is surprising that our method demonstrates such steady superiority}.

OC-VTP is designed as a train-once method, but it can also be trained individually on each benchmark that provides a training set. For fairness, we remove potential overlaps with the evaluation images. As MMB, MME, POPE, MMMU, and SEED are evaluation only, we do not train on these benchmarks. As expected, the performance on each benchmark yields additional improvement compared with our train-once OC-Pruner.

We also compare the results that do not pad the pruned tokens to the target budget $s$. When there are much less objects or object parts than \#slots, some slots can be empty, thus the retained \#tokens can be smaller than $s$. Across the ten benchmarks, the average number of retained tokens is 50.7 (8.8\%) and the average performance proportion is 92.6\%, which is even competitive with some SotA baselines that use the target number of tokens.

On VizWiz and SEED, pruned models can outperform the vanilla model. A possible reason is that redundant vision tokens somehow act as noise to image understanding.

\input{tables/llava-next-acc}

\subsubsection{Results on LLaVA-NeXT} OC-VTP achieves comparable results on LLaVA-NeXT with those on LLaVA-1.5 as shown in \Cref{tab:llava-next-acc}. When retaining only 5.6\% vision tokens, our pre-trained OC-VTP reduces the performance by merely 5.0\%. The performance gap between our method and the second-best SotA approach shows a consistent trend that at the 5.6\% retaining ratio, OC-VTP is 1.8\% better than VisionZip, but the gap narrows to 0.1\% and 1.0\% while retaining ratios increase, compared to HiPrune.
\textbf{Thus our method works better than the baselines in heavy pruning cases}.

On SQA, the results first decrease then increase with increasing pruning ratio.
At lower pruning ratios, tokens within the same cluster are pruned first, since these clusters represent the main object in the image and correspond to informative regions, performance initially drops. In this circumstance, the influence of noisy tokens becomes relatively stronger. While pruning continues, tokens from less important clusters (e.g., noise objects) are discarded, allowing the remaining representative object tokens to contribute more effectively, leading to a recovery of performance.

\input{tables/qwen2.5-vl}

\subsubsection{Results on Qwen2.5-VL}
Qwen2.5-VL uses a dynamic-resolution vision encoder with an adaptive token count \cite{qwen2.5-vl}, so we report the results by the retaining ratio rather than an exact number of retained tokens. As shown in \Cref{tab:qwen-acc}, the performance gap between OC-VTP and the second-best SotA method narrows faster. At ratio 33.3\%, OC-VTP is weaker than that of HiPrune. The overall performance on Qwen2.5-VL is also slightly weaker than on LLaVA VLMs. We believe that LLaVA produces a fixed number of tokens, so OC-Pruner can learn a stable pattern that maps a target budget to a fixed number of slots during training. In Qwen2.5-VL, the token count varies across images, so a train-once OC-Pruner can be misaligned with the actual budget, which weakens object-level feature aggregation and reduces pruning optimality.

\subsection{Results on Efficiency}
\subsubsection{Results on FLOPs}
We evaluate OC-VTP on LLaVA-1.5 and LLaVA-NeXT, and report the computation cost of the prefill FLOPs (MAC=2). The results are summarized in \Cref{tab:flops}.

\input{tables/flops}
\input{images_tex/efficiency_llava_time}
Compared with the vanilla model, our method effectively reduces computational overhead while maintaining performance. Specifically, assuming the prompt token length is 32, for LLaVA-1.5 the pruned model (retaining 64 tokens) requires only 0.97 T FLOPs compared to 6.30 T FLOPs in the original model, achieving a $6.5\times$ reduction. Similarly, for LLaVA-NeXT, the FLOPs drop from 33.76 T to 1.95 T, corresponding to a $17\times$ reduction. Meanwhile, our OC-Pruner only needs 5.97 G and 23.82 G FLOPs, which incurs nearly zero additional computational cost compared to the VLM backbone. Overall, the results demonstrate that the proposed OC-Pruner introduces only small overhead while substantially improving efficiency and preserving performance, making it highly scalable to large VLMs.

\subsubsection{Results on Latency}
We measure end-to-end inference latency per image on a single NVIDIA V100-32GB GPU with identical generation settings ($\text{batch size}=1$), using LLaVA-1.5 with 64 retained vision tokens and LLaVA-NeXT with 160 vision tokens. As shown in \Cref{fig:llava-time}, in GQA, MMB, POPE, VQA$^{\text{Text}}$ and SQA, pruning significantly reduces latency compared to vanilla VLMs. To better compare the effect of pruning, we also include a \textbf{Random pruner} baseline, which involves almost no extra overhead by randomly discarding vision tokens. On LLaVA-1.5 (left figure), the Random pruner and HiPrune achieve very similar latency and reduce the inference time per image by nearly 60\% relative to the vanilla model. Our OC-VTP is slightly slower but still comparable, yielding a 55\% (305.4 ms to 138.2 ms) reduction in both inference time and latency. On LLaVA-NeXT (right figure), pruning brings even greater benefits that HiPrune and Random pruner reduce the inference time by almost 70\%, while OC-VTP achieves a similar 65\% (811.8 ms to 287.3 ms) reduction. We expect these gains to be further improved when the pruners are deployed on more powerful GPUs, where computation dominates the system overhead.
\input{images_tex/visualization}
\subsection{Results on Visualization}
With a budget of 64 tokens on LLaVA-1.5, OC-Pruner retains one token per slot. As shown in \Cref{fig:visualization}, the selected tokens concentrate on different object centers (e.g., bike, cars, buildings, aircraft, animals, and signs). Large objects receive multiple slots that cover different parts, while small but important objects or background objects (such as the humans in (b) and (e), the mountain in the background in (c) and (f), and birds in (d)) still get at least one token, and backgrounds (sky, water, grass) are largely suppressed in several tokens unless they are important to define the scene. This object diversity and alignment match the expected behavior of OCL characteristics, explaining the reasons why OC-VTP preserving most of the performance. 

Typically, failure cases occur when a slot lands near, but not exactly on the instance. We believe this is because the objects are very small or have low contrast, making them difficult for the model to distinguish from the background. Another failure case appears in image (f), where the fence is omitted. We believe that this happens because when slot attention computes similarity with vision tokens, the fence is considered less important, and the slot instead selects tokens corresponding to the grass in the region.
\subsection{Ablations}
\input{tables/ablation_topk}
\textbf{Slot count and top-$k$ combinations.} 
We compare a different number of slots $s$ and the top-$k$ for each slot while keeping the same token budget $s \times k$. As shown in \Cref{tab:ablation_topk}, using smaller $k$ outperforms other combinations across all seven benchmarks. We therefore adopt $k=1$, which yields the best performance. In one slot cluster, similar tokens represent an object, so selecting more than one token per slot involves redundancy.
\input{tables/ablation_layer}

\textbf{Pruner insertion layers.} We compare OC-VTP at three insertion positions in the vision encoder that Layer 9, Layer 10, and Layer -2 under the same token budgets. Following the discoveries in HiPrune \cite{hiprune}, Layer 9 is recognized as the most \textbf{information-dense}, Layer 10 is the most \textbf{sparse}, and Layer -2 is near the encoder output. As shown in \Cref{tab:ablation_layer}, inserting at Layer 9 consistently yields the best results across budgets, so we adopt it as the first choice. Because Layer 9 is the most information-dense, it can represent the most objects with the fewest tokens. In very dense clusters, a single token is enough to represent the whole cluster.

\textbf{Loss functions.}
We compare AW-MSE with plain MSE on LLaVA-1.5, averaging the relative scores on seven benchmarks (GQA, MMB, MME, POPE, VQA$^{\text{Text}}$, MMMU, and SEED). As shown in \Cref{tab:ablation_loss}, training OC-Pruner with AW-MSE yields the highest accuracy in all token budgets, so we adopt AW-MSE as default. While the OCL module reconstructs encoder features, a plain MSE objective may underweight slots covering small but important regions. The inverse-area re-weighting in AW-MSE can solve this issue and preserve these informative areas.
\input{tables/ablation_loss}

%% file: tables/llava-1.5-acc.tex
\begin{table*}[h]
    \centering\small
    \setlength{\tabcolsep}{5pt}
    \resizebox{0.88\textwidth}{!}{
    \begin{tabular}{l|cccccccccc|c}
        \toprule
        Method & GQA & MMB & MME & POPE & SQA & VQA$^\text{Text}$ & VQA$^{v2}$ & VizWiz & MMMU & SEED & Avg \\
        \midrule
        \multicolumn{12}{c}{Vanilla, 576 Tokens (\textbf{100\%})} \\
        \multirow{2}{*}{Vanilla$^{\text{CVPR'24}}$} & 61.9 & 64.7 & 1862 & 85.9 & 69.5 & 58.2 & 78.5 & 55.8 & 36.3 & 58.6 & \multirow{2}{*}{100\%} \\
        & {\scriptsize 100\%} & {\scriptsize 100\%} & {\scriptsize 100\%} & {\scriptsize 100\%} & {\scriptsize 100\%} & {\scriptsize 100\%} & {\scriptsize 100\%} & {\scriptsize 100\%} & {\scriptsize 100\%} & {\scriptsize 100\%} \\
        \midrule
        \multicolumn{12}{c}{Retain 192 Tokens (\textbf{33.3\%})} \\
        \multirow{2}{*}{FastV$^{\text{ECCV'24}}$} & 52.7 & 61.2 & 1612 & 64.8 & 67.3 & 52.5 & 67.1 & 50.8 & 34.3 & 57.1 & \multirow{2}{*}{89.7\%} \\
        & {\scriptsize 85.1\%} & {\scriptsize 94.6\%} &{\scriptsize 86.6\%} & {\scriptsize 75.4\%} & {\scriptsize 96.8\%} & {\scriptsize 90.2\%} & {\scriptsize 85.5\%} & {\scriptsize 91.0\%} & {\scriptsize 94.5\%} & {\scriptsize 97.4\%} \\

        \midrule
        \multirow{2}{*}{SparseVLM$^{\text{ICML'25}}$} & 57.6 & 62.5 & 1721 & 83.6 & 69.1 & 56.1 & \textbf{77.0} & 50.6 & 33.8 & 55.8 & \multirow{2}{*}{95.2\%} \\
        & {\scriptsize 93.1\%} & {\scriptsize 96.6\%} & {\scriptsize 92.4\%} & {\scriptsize 97.3\%} & {\scriptsize 99.4\%} & {\scriptsize 96.4\%} & {\scriptsize \textbf{98.1\%}} & {\scriptsize 90.7\%} & {\scriptsize 93.1\%} & {\scriptsize 95.2\%} \\

        \midrule
        \multirow{2}{*}{VisionZip$^{\text{CVPR'25}}$} & 59.3 & 63.0 & 1783 & 85.3 & 68.9 & 57.3 & 76.8 & - & 36.6 & 56.4 & \multirow{2}{*}{97.9\%} \\
        & {\scriptsize 95.8\%} & {\scriptsize 97.4\%} & {\scriptsize 95.8\%} & {\scriptsize 99.3\%} & {\scriptsize 99.1\%} & {\scriptsize 98.5\%} & {\scriptsize 97.8\%} & {\scriptsize -} & {\scriptsize 101\%} & {\scriptsize 96.2\%} \\

        \midrule
        \multirow{2}{*}{PyramidDrop$^{\text{CVPR'25}}$} & 57.3 & \textbf{63.3} & 1797 & 82.3 & 69.0 & 56.5 & 75.1 & 51.1 & - & 54.7 & \multirow{2}{*}{95.5\%} \\
        & {\scriptsize 92.6\%} & {\scriptsize \textbf{97.8\%}} & {\scriptsize 96.5\%} & {\scriptsize 95.8\%} & {\scriptsize 99.3\%} & {\scriptsize 97.1\%} & {\scriptsize 95.7\%} & {\scriptsize 91.6\%} & {\scriptsize -} & {\scriptsize 93.3\%} \\    

        \midrule
        \multirow{2}{*}{HiPrune$^{\text{2025.08}}$} & 59.2 & 62.8 & \textbf{1814} & \textbf{86.1} & 68.9 & 57.6 & 76.7 & 54.5 & 36.4 & \textbf{63.6} & \multirow{2}{*}{99.3\%} \\
        & {\scriptsize 95.6\%} & {\scriptsize 97.1\%} & {\scriptsize \textbf{97.4\%}} & {\scriptsize \textbf{100\%}} & {\scriptsize 99.1\%} & {\scriptsize 99.0\%} & {\scriptsize 97.7\%} & {\scriptsize 97.7\%} & {\scriptsize 100\%} & {\scriptsize \textbf{109\%}} \\  

        \midrule
        \multirow{2}{*}{\textbf{OC-VTP$^{\text{Ours}}$}} & \textbf{59.4$\pm$0.4} & 63.1$\pm$0.3 & 1786$\pm$16 & 85.4$\pm$0.6 & \textbf{69.1$\pm$0.6} & \textbf{58.3$\pm$0.4} & 76.3$\pm$0.3 & \textbf{54.7$\pm$0.4} & \textbf{36.7$\pm$0.2} & 63.5$\pm$0.3 & \multirow{2}{*}{\textbf{99.3\%$\pm$0.3\%}} \\
        & {\scriptsize \textbf{95.9\%$\pm$0.6\%}} & {\scriptsize 97.5\%$\pm$0.4\%} & {\scriptsize 95.9\%$\pm$0.9\%} & {\scriptsize 99.4\%$\pm$0.8\%} & {\scriptsize \textbf{99.4\%$\pm$0.9\%}} & {\scriptsize \textbf{100.1\%$\pm$0.6\%}} & {\scriptsize 97.2\%$\pm$0.4\%} & {\scriptsize \textbf{98.0\%$\pm$0.7\%}} & {\scriptsize \textbf{101.0\%$\pm$0.6\%}} & {\scriptsize 108.3\%$\pm$0.6\%} \\ 

        \midrule
        \multirow{2}{*}{\textbf{\textcolor{gray}{OC-VTP$^{\text{Ours}}_{\blacktriangle}$}}} & \textcolor{gray}{59.6} & \textcolor{gray}{-} & \textcolor{gray}{-} & \textcolor{gray}{-} & \textcolor{gray}{69.0} & \textcolor{gray}{57.8} & \textcolor{gray}{77.4} & \textcolor{gray}{55.4} & \textcolor{gray}{-} & \textcolor{gray}{-} & \multirow{2}{*}{\textcolor{gray}{-}} \\
        & {\scriptsize \textcolor{gray}{96.3\%}} & {\scriptsize \textcolor{gray}{-}} & {\scriptsize \textcolor{gray}{-}} & {\scriptsize \textcolor{gray}{-}} & {\scriptsize \textcolor{gray}{99.3\%}} & {\scriptsize \textcolor{gray}{99.3\%}} & {\scriptsize \textcolor{gray}{98.6\%}} & {\scriptsize \textcolor{gray}{99.3\%}} & {\scriptsize \textcolor{gray}{-}} & {\scriptsize \textcolor{gray}{-}} \\ 

        \midrule
        \multicolumn{12}{c}{Retain 128 Tokens (\textbf{22.2\%})} \\
        \multirow{2}{*}{FastV$^{\text{ECCV'24}}$} & 49.6 & 56.1 & 1490 & 59.6 & 60.2 & 50.6 & 61.8 & 51.3 & 34.9 & 55.9 & \multirow{2}{*}{85.2\%} \\
        & {\scriptsize 80.1\%} & {\scriptsize 86.7\%} & {\scriptsize 80.0\%} & {\scriptsize 69.4\%} & {\scriptsize 86.6\%} & {\scriptsize 86.9\%} & {\scriptsize 78.7\%} & {\scriptsize 91.9\%} & {\scriptsize 96.1\%} & {\scriptsize 95.4\%} \\

        \midrule
        \multirow{2}{*}{SparseVLM$^{\text{ICML'25}}$} & 56.0 & 60.0 & 1696 & 80.5 & 67.1 & 54.9 & 73.8 & 50.2 & 33.8 & 53.4 & \multirow{2}{*}{92.7\%} \\
        & {\scriptsize 90.5\%} & {\scriptsize 92.7\%} & {\scriptsize 91.1\%} & {\scriptsize 93.7\%} & {\scriptsize 96.5\%} & {\scriptsize 94.3\%} & {\scriptsize 94.0\%} & {\scriptsize 90.0\%} & {\scriptsize 93.1\%} & {\scriptsize 91.1\%} \\

        \midrule
        \multirow{2}{*}{VisionZip$^{\text{CVPR'25}}$} & \textbf{57.6} & 62.0 & 1762 & 83.2 & 68.9 & 56.8 & \textbf{75.6} & - & \textbf{37.9} & 54.9 & \multirow{2}{*}{96.8\%} \\
        & {\scriptsize \textbf{93.1\%}} & {\scriptsize 95.8\%} & {\scriptsize 94.6\%} & {\scriptsize 96.9\%} & {\scriptsize 99.1\%} & {\scriptsize 97.6\%} & {\scriptsize \textbf{96.3\%}} & {\scriptsize -} & {\scriptsize \textbf{104\%}} & {\scriptsize 93.7\%} \\

        \midrule
        \multirow{2}{*}{PyramidDrop$^{\text{CVPR'25}}$} & 57.1 & 61.6 & 1761 & 82.3 & 68.4 & 56.6 & 72.9 & 51.0 & - & 53.3 & \multirow{2}{*}{94.3\%} \\
        & {\scriptsize 92.2\%} & {\scriptsize 95.2\%} & {\scriptsize 94.6\%} & {\scriptsize 95.8\%} & {\scriptsize 98.4\%} & {\scriptsize 97.3\%} & {\scriptsize 92.9\%} & {\scriptsize 91.4\%} & {\scriptsize -} & {\scriptsize 91.0\%} \\    

        \midrule
        \multirow{2}{*}{HiPrune$^{\text{2025.08}}$} & 57.3 & \textbf{62.2} & \textbf{1782} & 82.8 & 68.3 & 56.6 & 74.9 & 54.3 & 36.7 & \textbf{61.0} & \multirow{2}{*}{97.4\%} \\
        & {\scriptsize 92.6\%} & {\scriptsize \textbf{96.1\%}} & {\scriptsize \textbf{95.7\%}} & {\scriptsize 96.4\%} & {\scriptsize 98.3\%} & {\scriptsize 97.3\%} & {\scriptsize 95.4\%} & {\scriptsize 97.3\%} & {\scriptsize 101\%} & {\scriptsize \textbf{104\%}} \\  

        \midrule
        \multirow{2}{*}{\textbf{OC-VTP$^{\text{Ours}}$}} & 57.5$\pm$0.6 & 62.0$\pm$0.3 & 1742$\pm$14 & \textbf{84.4$\pm$0.6} & \textbf{69.7$\pm$0.5} & \textbf{56.8$\pm$0.4} & 74.9$\pm$0.1 & \textbf{55.1$\pm$0.4} & 36.3$\pm$0.1 & 60.3$\pm$1.0 & \multirow{2}{*}{\textbf{97.5\%$\pm$0.4\%}} \\
        & {\scriptsize 92.9\%$\pm$0.9\%} & {\scriptsize 95.8\%$\pm$0.4\%} & {\scriptsize 93.5\%$\pm$0.8\%} & {\scriptsize \textbf{98.3\%$\pm$0.8\%}} & {\scriptsize \textbf{100.3\%$\pm$0.8\%}} & {\scriptsize \textbf{97.6\%$\pm$0.7\%}} & {\scriptsize 95.5\%$\pm$0.2\%} & {\scriptsize \textbf{98.7\%$\pm$0.7\%}} & {\scriptsize 99.9\%$\pm$0.3\%} & {\scriptsize 103.0\%$\pm$1.7\%} \\ 

        \midrule
        \multirow{2}{*}{\textbf{\textcolor{gray}{OC-VTP$^{\text{Ours}}_{\blacktriangle}$}}} & \textcolor{gray}{58.4} & \textcolor{gray}{-} & \textcolor{gray}{-} & \textcolor{gray}{-} & \textcolor{gray}{69.0} & \textcolor{gray}{56.9} & \textcolor{gray}{76.2} & \textcolor{gray}{54.3} & \textcolor{gray}{-} & \textcolor{gray}{-} & \multirow{2}{*}{\textcolor{gray}{-}} \\
        & {\scriptsize \textcolor{gray}{94.3\%}} & {\scriptsize \textcolor{gray}{-}} & {\scriptsize \textcolor{gray}{-}} & {\scriptsize \textcolor{gray}{-}} & {\scriptsize \textcolor{gray}{99.3\%}} & {\scriptsize \textcolor{gray}{97.8\%}} & {\scriptsize \textcolor{gray}{97.1\%}} & {\scriptsize \textcolor{gray}{97.3\%}} & {\scriptsize \textcolor{gray}{-}} & {\scriptsize \textcolor{gray}{-}} \\ 

        \midrule
        \multicolumn{12}{c}{Retain 64 Tokens (\textbf{11.1\%})} \\
        \multirow{2}{*}{FastV$^{\text{ECCV'24}}$} & 46.1 & 48.1 & 1256 & 48.0 & 51.1 & 47.8 & 55.0 & 50.8 & 34.0 & 51.9 & \multirow{2}{*}{77.1\%} \\
        & {\scriptsize 74.5\%} & {\scriptsize 74.3\%} & {\scriptsize 67.5\%} & {\scriptsize 55.9\%} & {\scriptsize 73.5\%} & {\scriptsize 82.1\%} & {\scriptsize 70.1\%} & {\scriptsize 91.0\%} & {\scriptsize 93.7\%} & {\scriptsize 88.6\%} \\

        \midrule
        \multirow{2}{*}{SparseVLM$^{\text{ICML'25}}$} & 52.7 & 56.2 & 1505 & 75.1 & 62.2 & 51.8 & 68.2 & 50.4 & 32.7 & 51.1 & \multirow{2}{*}{87.3\%} \\
        & {\scriptsize 85.1\%} & {\scriptsize 86.9\%} & {\scriptsize 80.8\%} & {\scriptsize 87.4\%} & {\scriptsize 89.5\%} & {\scriptsize 89.0\%} & {\scriptsize 86.9\%} & {\scriptsize 90.3\%} & {\scriptsize 90.1\%} & {\scriptsize 87.2\%} \\

        \midrule
        \multirow{2}{*}{VisionZip$^{\text{CVPR'25}}$} & \textbf{55.1} & \textbf{60.1} & \textbf{1690} & 77.0 & 69.0 & \textbf{55.5} & 72.4 & - & 36.2 & 52.2 & \multirow{2}{*}{93.1\%} \\
        & {\scriptsize \textbf{89.0\%}} & {\scriptsize \textbf{92.9\%}} & {\scriptsize \textbf{90.8\%}} & {\scriptsize 89.6\%} & {\scriptsize 99.3\%} & {\scriptsize \textbf{95.4\%}} & {\scriptsize 92.2\%} & {\scriptsize -} & {\scriptsize 99.7\%} & {\scriptsize 89.1\%} \\

        \midrule
        \multirow{2}{*}{PyramidDrop$^{\text{CVPR'25}}$} & 47.5 & 58.8 & 1561 & 55.9 & 69.0 & 50.6 & 69.2 & 50.7 & - & 50.4 & \multirow{2}{*}{85.3\%} \\
        & {\scriptsize 76.7\%} & {\scriptsize 90.9\%} & {\scriptsize 83.8\%} & {\scriptsize 65.1\%} & {\scriptsize 99.3\%} & {\scriptsize 86.9\%} & {\scriptsize 88.2\%} & {\scriptsize 90.9\%} & {\scriptsize -} & {\scriptsize 86.0\%} \\    

        \midrule
        \multirow{2}{*}{HiPrune$^{\text{2025.08}}$} & 53.6 & 59.5 & 1646 & 73.0 & 68.9 & 54.9 & 69.2 & 54.4 & \textbf{36.7} & 55.2 & \multirow{2}{*}{92.6\%} \\
        & {\scriptsize 86.6\%} & {\scriptsize 92.0\%} & {\scriptsize 88.4\%} & {\scriptsize 85.0\%} & {\scriptsize 99.1\%} & {\scriptsize 94.3\%} & {\scriptsize 88.2\%} & {\scriptsize 97.5\%} & {\scriptsize \textbf{101\%}} & {\scriptsize 94.2\%} \\  

        \midrule
        \multirow{2}{*}{\textbf{OC-VTP$^{\text{Ours}}$}} 
        & 54.6$\pm$1.0 & 59.3$\pm$0.8 & 1668$\pm$28 & \textbf{80.2$\pm$2.2} & \textbf{69.4$\pm$0.5} & 55.0$\pm$0.3 & \textbf{73.2$\pm$0.3} & \textbf{55.4$\pm$1.0} & 36.0$\pm$0.3 & \textbf{56.4$\pm$1.0} & \multirow{2}{*}{\textbf{94.5\%$\pm$0.9\%}} \\
        & {\scriptsize 88.1\%$\pm$1.7\%} & {\scriptsize 91.6\%$\pm$1.2\%} & {\scriptsize 89.6\%$\pm$1.5\%} & {\scriptsize \textbf{93.3\%$\pm$2.5\%}} & {\scriptsize \textbf{99.8\%$\pm$0.8\%}} & {\scriptsize 94.5\%$\pm$0.6\%} & {\scriptsize \textbf{93.3\%$\pm$0.4\%}} & {\scriptsize \textbf{99.3\%$\pm$1.7\%}} & {\scriptsize 99.2\%$\pm$0.8\%} & {\scriptsize \textbf{96.3\%$\pm$1.8\%}} \\ 

        \midrule
        \multirow{2}{*}{\textbf{\textcolor{gray}{OC-VTP$^{\text{Ours}}_{\blacktriangle}$}}} & \textcolor{gray}{56.2} & \textcolor{gray}{-} & \textcolor{gray}{-} & \textcolor{gray}{-} & \textcolor{gray}{69.0} & \textcolor{gray}{55.4} & \textcolor{gray}{75.1} & \textcolor{gray}{55.2} & \textcolor{gray}{-} & \textcolor{gray}{-} & \multirow{2}{*}{\textcolor{gray}{-}} \\
        & {\scriptsize \textcolor{gray}{90.8\%}} & {\scriptsize \textcolor{gray}{-}} & {\scriptsize \textcolor{gray}{-}} & {\scriptsize \textcolor{gray}{-}} & {\scriptsize \textcolor{gray}{99.3\%}} & {\scriptsize \textcolor{gray}{95.2\%}} & {\scriptsize \textcolor{gray}{95.7\%}} & {\scriptsize \textcolor{gray}{98.9\%}} & {\scriptsize \textcolor{gray}{-}} & {\scriptsize \textcolor{gray}{-}} \\ 

        \midrule
        \multicolumn{12}{c}{Retain 50.7 Tokens (\textbf{8.8\%})} \\
        \multirow{2}{*}{\textbf{OC-VTP$^{\text{Ours}}$}} & 55.3 & 56.8 & 1581 & 81.5 & 68.6 & 51.5 & 67.4 & 56.6 & 35.4 & 57.1 & \multirow{2}{*}{\textbf{92.6\%}} \\
        & {\scriptsize 89.3\%} & {\scriptsize 87.8\%} & {\scriptsize 84.9\%} & {\scriptsize 94.9\%} & {\scriptsize 98.7\%} & {\scriptsize 88.5\%} & {\scriptsize 85.9\%} & {\scriptsize 101\%} & {\scriptsize 97.5\%} & {\scriptsize 97.4\%} \\
        \bottomrule
    \end{tabular}}
    \caption{
    Performance comparison on LLaVA-1.5. OC-VTP is evaluated over five seeds.
    The vanilla number of vision tokens is 576. The first line of each method shows the accuracy, and the second line shows the relative accuracy to the vanilla score. The last column is the average proportion across all benchmarks. $\blacktriangle$ means that OC-VTP is trained on each individual benchmark's training set, with no overlap with the scoring splits. "-" means the benchmark does not have a training set or a published result. The 8.8\% retained tokens refer to the average number of remaining tokens across all benchmarks if the pruned tokens are not padded to 64 tokens.
    }
    \label{tab:llava-1.5-acc}
\end{table*}

%% file: tables/llava-next-acc.tex

\begin{table}[]
    \centering\small
    \resizebox{0.48\textwidth}{!}{
    \begin{tabular}{l|cccccc|c}
        \toprule
        Method & GQA & MMB & POPE & SQA & VQA$^\text{Text}$ & VizWiz & Avg \\
        \midrule
        \multicolumn{8}{c}{Vanilla, 2880 Tokens (\textbf{100\%})} \\
        \multirow{2}{*}{Vanilla$^{\text{CVPR'24}}$} & 64.2 & 66.4 & 86.2 & 67.5 & 61.3 & 55.2 & \multirow{2}{*}{100\%} \\
        & {\scriptsize 100\%} & {\scriptsize 100\%} & {\scriptsize 100\%} & {\scriptsize 100\%} & {\scriptsize 100\%} & {\scriptsize 100\%}\\
        \midrule
        \multicolumn{8}{c}{Retain 640 Tokens (\textbf{22.2\%})} \\

        \multirow{2}{*}{VisionZip$^{\text{CVPR'25}}$} & \textbf{61.3} & 66.3 & 86.3 & 68.1 & \textbf{60.2} & 57.1 & \multirow{2}{*}{99.7\%} \\
        & {\scriptsize \textbf{95.5\%}} & {\scriptsize 99.8\%} & {\scriptsize 100\%} & {\scriptsize 101\%} & {\scriptsize \textbf{98.2\%}} & {\scriptsize 103\%} \\ 

        \midrule
        \multirow{2}{*}{HiPrune$^{\text{2025.08}}$} & 60.6 & \textbf{67.0} & 85.3 & 68.0 & 60.0 & 59.9 & \multirow{2}{*}{100\%} \\
        & {\scriptsize 94.4\%} & {\scriptsize \textbf{101\%}} & {\scriptsize 99.0\%} & {\scriptsize 101\%} & {\scriptsize 97.9\%} & {\scriptsize 109\%} \\  

        \midrule
        \multirow{2}{*}{\textbf{OC-VTP$^{\text{Ours}}$}} & 61.2 & 64.8 & \textbf{86.9} & \textbf{69.3} & 60.1 & \textbf{60.3} & \multirow{2}{*}{\textbf{101\%}} \\
        & {\scriptsize 95.3\%} & {\scriptsize 97.6\%} & {\scriptsize 101\%} & {\scriptsize 103\%} & {\scriptsize 98.0\%} & {\scriptsize 109\%} \\ 

        \midrule
        \multirow{2}{*}{\textbf{\textcolor{gray}{OC-VTP$^{\text{Ours}}_{\blacktriangle}$}}} & \textcolor{gray}{62.8} & \textcolor{gray}{-} & \textcolor{gray}{-} & \textcolor{gray}{69.3} & \textcolor{gray}{61.1} & \textcolor{gray}{60.1} & \multirow{2}{*}{\textcolor{gray}{-}} \\
        & {\scriptsize \textcolor{gray}{97.8\%}} & {\scriptsize \textcolor{gray}{-}} & {\scriptsize \textcolor{gray}{-}} & {\scriptsize \textcolor{gray}{103\%}} & {\scriptsize \textcolor{gray}{99.7\%}} & {\scriptsize \textcolor{gray}{109\%}}  \\ 

        \midrule
        \multicolumn{8}{c}{Retain 320 Tokens (\textbf{11.1\%})} \\

        \multirow{2}{*}{VisionZip$^{\text{CVPR'25}}$} & \textbf{59.3} & 63.1 & 82.1 & 67.3 & \textbf{58.9} & 56.2 & \multirow{2}{*}{96.7\%} \\
        & {\scriptsize \textbf{92.4\%}} & {\scriptsize 95.0\%} & {\scriptsize 95.2\%} & {\scriptsize 99.7\%} & {\scriptsize \textbf{96.1\%}} & {\scriptsize 102\%} \\

        \midrule
        \multirow{2}{*}{HiPrune$^{\text{2025.08}}$} & 57.4 & \textbf{65.3} & 78.9 & 67.3 & 57.6 & 59.9 & \multirow{2}{*}{96.9\%} \\
        & {\scriptsize 89.4\%} & {\scriptsize \textbf{98.3\%}} & {\scriptsize 91.5\%} & {\scriptsize 99.7\%} & {\scriptsize 94.0\%} & {\scriptsize 109\%} \\  

        \midrule
        \multirow{2}{*}{\textbf{OC-VTP$^{\text{Ours}}$}} & 58.1 & 61.7 & \textbf{82.3} & \textbf{67.4} & 58.2 & \textbf{59.9} & \multirow{2}{*}{\textbf{97.0\%}} \\
        & {\scriptsize 90.5\%} & {\scriptsize 92.9\%} & {\scriptsize \textbf{95.5\%}} & {\scriptsize \textbf{99.9\%}} & {\scriptsize 94.9\%} & {\scriptsize \textbf{109\%}} \\ 

        \midrule
        \multirow{2}{*}{\textbf{\textcolor{gray}{OC-VTP$^{\text{Ours}}_{\blacktriangle}$}}} & \textcolor{gray}{59.6} & \textcolor{gray}{-} & \textcolor{gray}{-} & \textcolor{gray}{67.3} & \textcolor{gray}{58.4} & \textcolor{gray}{59.6} & \multirow{2}{*}{\textcolor{gray}{-}} \\
        & {\scriptsize \textcolor{gray}{92.8\%}} & {\scriptsize \textcolor{gray}{-}} & {\scriptsize \textcolor{gray}{-}} & {\scriptsize \textcolor{gray}{99.7\%}} & {\scriptsize \textcolor{gray}{95.3\%}} & {\scriptsize \textcolor{gray}{108\%}} \\ 

        \midrule
        \multicolumn{8}{c}{Retain 160 Tokens (\textbf{5.6\%})} \\

        \multirow{2}{*}{VisionZip$^{\text{CVPR'25}}$} & 55.1 & \textbf{60.1} & 77.0 & 69.0 & 55.5 & 55.5 & \multirow{2}{*}{93.1\%} \\
        & {\scriptsize 85.8\%} & {\scriptsize \textbf{90.5\%}} & {\scriptsize 89.3\%} & {\scriptsize 102\%} & {\scriptsize 90.5\%} & {\scriptsize 101\%} \\  

        \midrule
        \multirow{2}{*}{HiPrune$^{\text{2025.08}}$} & 52.5 & 59.8 & 67.7 & 68.7 & 51.2 & 57.2 & \multirow{2}{*}{89.9\%} \\
        & {\scriptsize 81.8\%} & {\scriptsize 90.1\%} & {\scriptsize 78.5\%} & {\scriptsize 102\%} & {\scriptsize 83.5\%} & {\scriptsize 104\%} \\  
        \midrule
        \multirow{2}{*}{\textbf{OC-VTP$^{\text{Ours}}$}} & \textbf{56.1} & 60.0 & \textbf{78.9} & \textbf{69.2} & \textbf{55.7} & \textbf{59.0} & \multirow{2}{*}{\textbf{94.9\%}} \\
        & {\scriptsize \textbf{87.4\%}} & {\scriptsize 90.4\%} & {\scriptsize \textbf{91.5\%}} & {\scriptsize \textbf{103\%}} & {\scriptsize \textbf{90.9\%}} & {\scriptsize \textbf{107\%}} \\ 

        \midrule
        \multirow{2}{*}{\textbf{\textcolor{gray}{OC-VTP$^{\text{Ours}}_{\blacktriangle}$}}} & \textcolor{gray}{57.6} & \textcolor{gray}{-} & \textcolor{gray}{-} & \textcolor{gray}{69.0} & \textcolor{gray}{56.8} & \textcolor{gray}{59.2} & \multirow{2}{*}{\textcolor{gray}{-}} \\
        & {\scriptsize \textcolor{gray}{89.7\%}} & {\scriptsize \textcolor{gray}{-}} & {\scriptsize \textcolor{gray}{-}} & {\scriptsize \textcolor{gray}{102\%}} & {\scriptsize \textcolor{gray}{92.7\%}} & {\scriptsize \textcolor{gray}{107\%}} \\ 
        \bottomrule
    \end{tabular}
    }
    \caption{
    Performance of OC-VTP on LLaVA-NeXT.
    The vanilla number of vision tokens is 2880. $\blacktriangle$ means that OC-VTP is trained on each individual benchmark's training set, ensuring no overlap with the scoring splits. "-" means the benchmark does not have a training set for individual training.}
    \label{tab:llava-next-acc}
\end{table}

%% file: tables/qwen2.5-vl.tex

\begin{table}[]
    \centering\small
    \resizebox{0.4\textwidth}{!}{
    \begin{tabular}{l|cccc|c}
        \toprule
        Method & MMB & POPE & SQA & VizWiz & Avg \\
        \midrule
        \multicolumn{6}{c}{Vanilla, \textbf{100\%} Tokens} \\
        \multirow{2}{*}{Vanilla$^{\text{CVPR'24}}$} & 77.3 & 87.0 & 80.4 & 68.3 & \multirow{2}{*}{100\%} \\
        & {\scriptsize 100\%} & {\scriptsize 100\%} & {\scriptsize 100\%} & {\scriptsize 100\%}\\
        \midrule
        \multicolumn{6}{c}{Retain \textbf{33.3\%} Tokens} \\

        \multirow{2}{*}{VisionZip$^{\text{CVPR'25}}$} & 74.9 & 85.4 & \textbf{80.1} & 67.1 & \multirow{2}{*}{98.2\%} \\
        & {\scriptsize 96.9\%} & {\scriptsize 98.2\%} & {\scriptsize \textbf{99.6\%}} & {\scriptsize 98.2\%} \\ 

        \midrule
        \multirow{2}{*}{HiPrune$^{\text{2025.08}}$} & \textbf{75.8} & 86.0 & 80.0 & \textbf{67.5} & \multirow{2}{*}{\textbf{98.8\%}} \\
        & {\scriptsize \textbf{98.1\%}} & {\scriptsize 98.9\%} & {\scriptsize 99.5\%} & {\scriptsize \textbf{98.8\%}} \\  

        \midrule
        \multirow{2}{*}{\textbf{OC-VTP$^{\text{Ours}}$}} & 75.3 & \textbf{86.4} & 79.6 & 67.4 & \multirow{2}{*}{98.6\%} \\
        & {\scriptsize 97.4\%} & {\scriptsize \textbf{99.3\%}} & {\scriptsize 99.0\%} & {\scriptsize 98.7\%} \\ 

        \midrule
        \multirow{2}{*}{\textbf{\textcolor{gray}{OC-VTP$^{\text{Ours}}_{\blacktriangle}$}}} & \textcolor{gray}{-} & \textcolor{gray}{-} & \textcolor{gray}{79.8} & \textcolor{gray}{67.5} & \multirow{2}{*}{\textcolor{gray}{-}} \\
        & {\scriptsize \textcolor{gray}{-}} & {\scriptsize \textcolor{gray}{-}} & {\scriptsize \textcolor{gray}{99.3\%}} & {\scriptsize \textcolor{gray}{98.8\%}}  \\ 

        \midrule
        \multicolumn{6}{c}{Retain \textbf{22.2\%} Tokens} \\

        \multirow{2}{*}{VisionZip$^{\text{CVPR'25}}$} & 73.5 & 84.6 & 80.0 & 66.3 & \multirow{2}{*}{97.2\%} \\
        & {\scriptsize 95.1\%} & {\scriptsize 97.2\%} & {\scriptsize 99.5\%} & {\scriptsize 97.1\%} \\

        \midrule
        \multirow{2}{*}{HiPrune$^{\text{2025.08}}$} & 74.0 & 84.7 & \textbf{80.3} & 66.5 & \multirow{2}{*}{97.6\%} \\
        & {\scriptsize 95.7\%} & {\scriptsize 97.4\%} & {\scriptsize \textbf{99.9\%}} & {\scriptsize 97.4\%} \\  

        \midrule
        \multirow{2}{*}{\textbf{OC-VTP$^{\text{Ours}}$}} & \textbf{74.6} & \textbf{85.0} & 79.8 & \textbf{66.8} & \multirow{2}{*}{\textbf{97.8\%}} \\
        & {\scriptsize \textbf{96.5\%}} & {\scriptsize \textbf{97.7\%}} & {\scriptsize 99.3\%} & {\scriptsize \textbf{97.8\%}} \\ 

        \midrule
        \multirow{2}{*}{\textbf{\textcolor{gray}{OC-VTP$^{\text{Ours}}_{\blacktriangle}$}}} & \textcolor{gray}{-} & \textcolor{gray}{-} & \textcolor{gray}{80.0} & \textcolor{gray}{67.1} & \multirow{2}{*}{\textcolor{gray}{-}} \\
        & {\scriptsize \textcolor{gray}{-}} & {\scriptsize \textcolor{gray}{-}} & {\scriptsize \textcolor{gray}{99.5\%}} & {\scriptsize \textcolor{gray}{98.2\%}} \\ 

        \midrule
        \multicolumn{6}{c}{Retain \textbf{11.1\%} Tokens} \\

        \multirow{2}{*}{VisionZip$^{\text{CVPR'25}}$} & 67.8 & 80.2 & \textbf{79.5} & 62.8 & \multirow{2}{*}{92.7\%} \\
        & {\scriptsize 87.7\%} & {\scriptsize 92.2\%} & {\scriptsize \textbf{98.9\%}} & {\scriptsize 91.9\%} \\  

        \midrule
        \multirow{2}{*}{HiPrune$^{\text{2025.08}}$} & 69.6 & 80.4 & 78.9 & 64.2 & \multirow{2}{*}{93.6\%} \\
        & {\scriptsize 90.0\%} & {\scriptsize 92.4\%} & {\scriptsize 98.1\%} & {\scriptsize 94.0\%} \\  

        \midrule
        \multirow{2}{*}{\textbf{OC-VTP$^{\text{Ours}}$}} & \textbf{70.9} & \textbf{81.8} & 78.0 & \textbf{64.2} & \multirow{2}{*}{\textbf{94.2\%}} \\
        & {\scriptsize \textbf{91.7\%}} & {\scriptsize \textbf{94.0\%}} & {\scriptsize 97.0\%} & {\scriptsize \textbf{94.0\%}} \\ 

        \midrule
        \multirow{2}{*}{\textbf{\textcolor{gray}{OC-VTP$^{\text{Ours}}_{\blacktriangle}$}}} & \textcolor{gray}{-} & \textcolor{gray}{-} & \textcolor{gray}{78.2} & \textcolor{gray}{66.1} & \multirow{2}{*}{\textcolor{gray}{-}} \\
        & {\scriptsize \textcolor{gray}{-}} & {\scriptsize \textcolor{gray}{-}} & {\scriptsize \textcolor{gray}{97.3\%}} & {\scriptsize \textcolor{gray}{96.8\%}} \\ 
        \bottomrule
    \end{tabular}
    }
    \caption{
    Performance of OC-VTP on Qwen2.5-VL.
    Qwen2.5-VL uses adaptive token counts, retaining ratio is used for evaluation. $\blacktriangle$ means that OC-VTP is trained on each individual benchmark's training set, ensuring no overlap with the scoring splits. "-" means the benchmark does not have a training set for individual training.}
    \label{tab:qwen-acc}
\end{table}

%% file: tables/flops.tex
\begin{table}
\small
\centering
\begin{tabular}{@{}lccc@{}}
    \toprule
    Method & Vanilla & Pruned & OC-Pruner \\
    \midrule
    LLaVA-1.5 & 6.30 T & 0.97 T & 5.97 G \\
    LLaVA-NeXT & 33.76 T & 1.95 T & 23.82 G \\
    \bottomrule
\end{tabular}
\caption{
Prefill FLOPs (MAC=2).
Computed per image for the prefill stage with 32 text tokens. "Vanilla" is the unpruned VLM, "Pruned" is the VLM after token pruning (64 vision tokens for LLaVA-1.5, and 160 for LLaVA-NeXT), and "OC-Pruner" is the additional cost of our pruner. T = teraFLOPs, G = gigaFLOPs.
}
\label{tab:flops}
\end{table}

%% file: images_tex/efficiency_llava_time.tex


\begin{figure}
  \centering
  \begin{subfigure}{0.48\linewidth}
    \includegraphics[width=\linewidth]{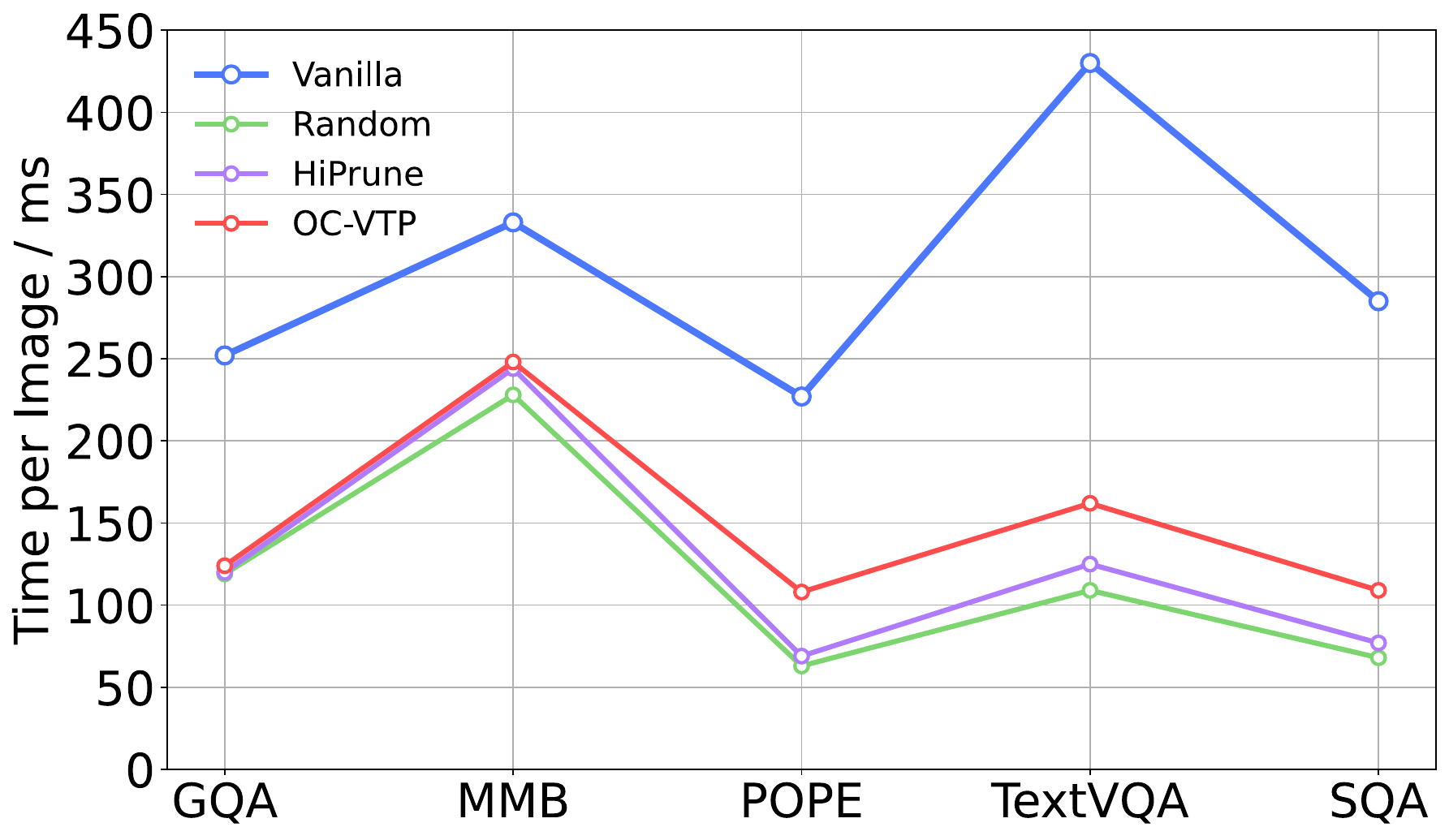}
    \caption{\scriptsize
    Inference time per image of LLaVA-1.5 with 64 retrained vision tokens.
    }
    \label{fig:short-a}
  \end{subfigure}
  \hfill
  \begin{subfigure}{0.48\linewidth}
    \includegraphics[width=\linewidth]{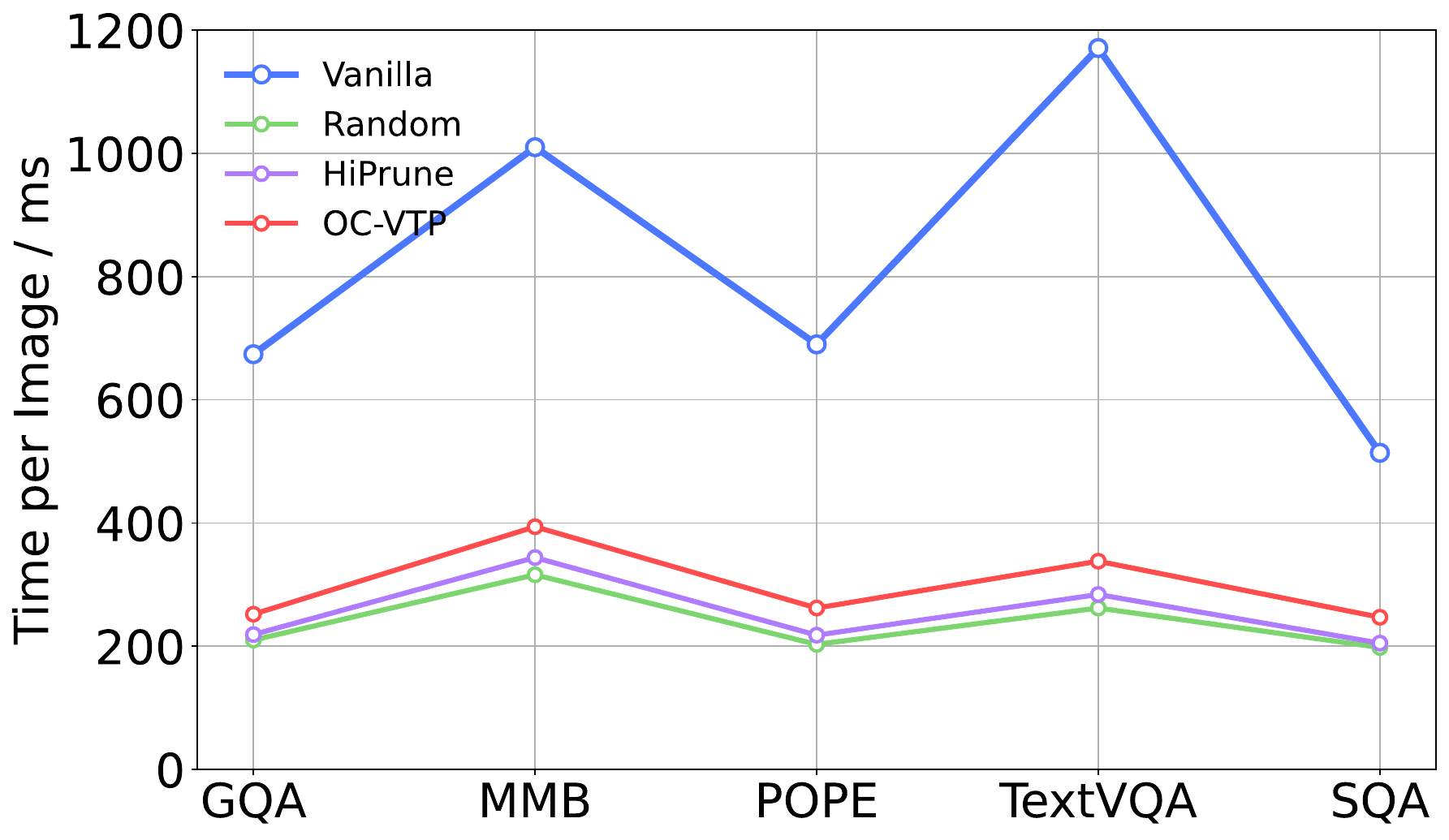}
    \caption{\scriptsize
    Inference time per image of LLaVA-NeXT with 160 tokens.
    }
    \label{fig:short-b}
  \end{subfigure}
  \caption{
  Inference time per image (ms).
  Average inference time for each case/image is calculated from total evaluation using LMMs-Eval on a single V100-32G GPU with $\text{batch size}=1$. Notably, \textit{Random} is a random pruning baseline with no extra cost.
  }
  \label{fig:llava-time}
\end{figure}

%% file: images_tex/visualization.tex
\begin{figure*}
  \centering
  \begin{subfigure}{0.16\linewidth}
    \includegraphics[width=\linewidth]{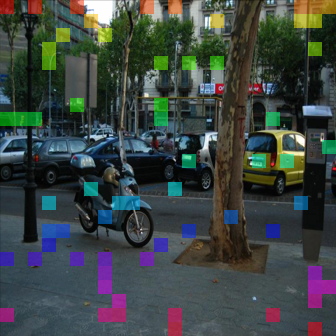}
    \caption{\scriptsize
    Tokens from the bike, cars, ground, trees, and buildings.
    }
    \label{fig:short-a}
  \end{subfigure}
  \hfill
  \begin{subfigure}{0.16\linewidth}
    \includegraphics[width=\linewidth]{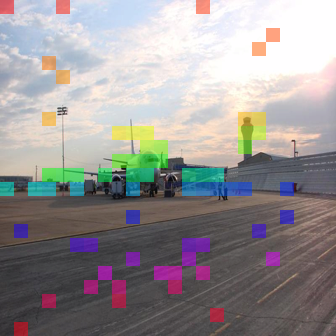}
    \caption{\scriptsize
    Tokens from the aircraft, humans, the tower, clouds, and the runway.
    }
    \label{fig:short-b}
  \end{subfigure}
  \hfill
  \begin{subfigure}{0.16\linewidth}
    \includegraphics[width=\linewidth]{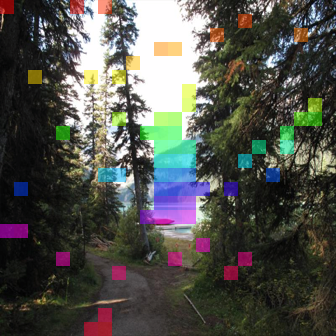}
    \caption{\scriptsize
    Tokens from trees, the mountain, the sky, and the path.
    }
    \label{fig:short-c}
  \end{subfigure}
  \hfill
  \begin{subfigure}{0.16\linewidth}
    \includegraphics[width=\linewidth]{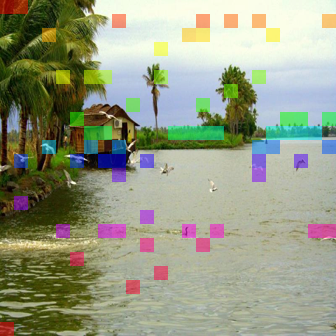}
    \caption{\scriptsize
    Tokens from the house, trees, birds, the sky, clouds and the lake.
    }
    \label{fig:short-d}
  \end{subfigure}
  \hfill
  \begin{subfigure}{0.16\linewidth}
    \includegraphics[width=\linewidth]{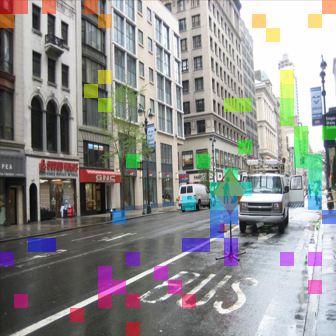}
    \caption{\scriptsize
    Tokens from signs, buildings, the human, the minivan, and lanes.
    }
    \label{fig:short-e}
  \end{subfigure}
  \hfill
  \begin{subfigure}{0.16\linewidth}
    \includegraphics[width=\linewidth]{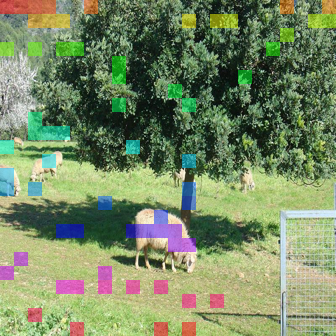}
    \caption{\scriptsize
    Tokens from sheep, grasses, the trunk, leaves and the mountain.
    }
    \label{fig:short-f}
  \end{subfigure}
  \caption{
  \textbf{OC-VTP visualization results.} OC-Pruner retains one token each slot, and the slots represent different objects in the scene. The retaining budget is 64, and the test is conducted on LLaVA-1.5.
  }
  \label{fig:visualization}
\end{figure*}

%% file: tables/ablation_topk.tex
\begin{table}
    \centering
    \resizebox{0.48\textwidth}{!}{
    \begin{tabular}{l|ccccccc|c}
        \toprule
        slots top-$k$ & GQA & MMB & MME & POPE & VQA$^\text{Text}$ & MMMU & SEED & Avg \\
        \midrule
        \multicolumn{9}{c}{Vanilla, 576 Tokens (\textbf{100\%})} \\
        \multirow{2}{*}{Vanilla} & 61.9 & 64.7 & 1862 & 85.9 & 58.2 & 36.3 & 58.6 & \multirow{2}{*}{100\%} \\
        & {\scriptsize 100\%} & {\scriptsize 100\%} & {\scriptsize 100\%} & {\scriptsize 100\%} & {\scriptsize 100\%} & {\scriptsize 100\%} & {\scriptsize 100\%} \\
        \midrule
        \multicolumn{9}{c}{Retain 192 Tokens (\textbf{33.3\%})} \\

        \multirow{2}{*}{32 Slots - Top 6} & 56.4 & 60.9 & 1712 & 83.9 & 55.2 & 33.6 & 58.8 & \multirow{2}{*}{94.6\%} \\
        & {\scriptsize 91.1\%} & {\scriptsize 94.1\%} & {\scriptsize 91.9\%} & {\scriptsize 97.7\%} & {\scriptsize 94.8\%} & {\scriptsize 92.6\%} & {\scriptsize 100\%}  \\ 

        \midrule
        \multirow{2}{*}{64 Slots - Top 3} & 58.1 & 61.5 & 1738 & 85.6 & 55.6 & 35.2 & 60.0 & \multirow{2}{*}{96.7\%} \\
        & {\scriptsize 93.9\%} & {\scriptsize 95.1\%} & {\scriptsize 93.3\%} & {\scriptsize 99.7\%} & {\scriptsize 95.5\%} & {\scriptsize 97.0\%} & {\scriptsize 102\%}  \\ 

        \midrule
        \multirow{2}{*}{\textbf{192 Slots - Top 1}} & \textbf{59.5} & \textbf{63.4} & \textbf{1808} & \textbf{86.3} & \textbf{57.8} & \textbf{36.4} & \textbf{63.4} & \multirow{2}{*}{\textbf{99.9\%}} \\
        & {\scriptsize \textbf{96.1\%}} & {\scriptsize \textbf{98.0\%}} & {\scriptsize \textbf{97.1\%}} & {\scriptsize \textbf{101\%}} & {\scriptsize \textbf{99.3\%}} & {\scriptsize \textbf{100\%}} & {\scriptsize \textbf{108\%}}  \\ 

        \midrule
        \multicolumn{9}{c}{Retain 128 Tokens (\textbf{22.2\%})} \\  

        \multirow{2}{*}{32 Slots - Top 4} & 55.2 & 58.1 & 1678 & 82.8 & 54.3 & 35.9 & 54.0 & \multirow{2}{*}{92.8\%} \\
        & {\scriptsize 89.2\%} & {\scriptsize 89.8\%} & {\scriptsize 90.1\%} & {\scriptsize 96.4\%} & {\scriptsize 93.3\%} & {\scriptsize 98.9\%} & {\scriptsize 92.2\%} \\ 
        \midrule

        \multirow{2}{*}{64 Slots - Top 2} & 56.9 & 61.3 & 1729 & 84.0 & 54.9 & 35.8 & 59.1 & \multirow{2}{*}{95.9\%} \\
        & {\scriptsize 91.9\%} & {\scriptsize 94.7\%} & {\scriptsize 92.9\%} & {\scriptsize 97.8\%} & {\scriptsize 94.3\%} & {\scriptsize 98.6\%} & {\scriptsize 101\%} \\ 
        \midrule

        \multirow{2}{*}{\textbf{128 Slots - Top 1}} & \textbf{58.0} & \textbf{62.2} & \textbf{1761} & \textbf{85.3} & \textbf{55.6} & \textbf{36.1} & \textbf{60.9} & \multirow{2}{*}{\textbf{97.5\%}} \\
        & {\scriptsize \textbf{93.7\%}} & {\scriptsize \textbf{96.1\%}} & {\scriptsize \textbf{94.6\%}} & {\scriptsize \textbf{99.3\%}} & {\scriptsize \textbf{95.5\%}} & {\scriptsize \textbf{99.4\%}} & {\scriptsize \textbf{104\%}}  \\ 
        
        \multicolumn{9}{c}{Retain 64 Tokens (\textbf{11.1\%})} \\
    
        \multirow{2}{*}{32 Slots - Top 2} & 53.6 & 57.8 & 1612 & 81.7 & 52.9 & 35.3 & 54.0 & \multirow{2}{*}{91.1\%} \\
        & {\scriptsize 86.6\%} & {\scriptsize 89.3\%} & {\scriptsize 86.6\%} & {\scriptsize 95.1\%} & {\scriptsize 90.9\%} & {\scriptsize 97.2\%} & {\scriptsize 92.2\%} \\ 

        \midrule
        \multirow{2}{*}{\textbf{64 Slots - Top 1}} & \textbf{55.7} & \textbf{59.8} & \textbf{1708} & \textbf{83.0} & \textbf{54.6} & \textbf{35.6} & \textbf{57.9} & \multirow{2}{*}{\textbf{94.5\%}} \\
        & {\scriptsize \textbf{90.0\%}} & {\scriptsize \textbf{92.4\%}} & {\scriptsize \textbf{91.7\%}} & {\scriptsize \textbf{96.6\%}} & {\scriptsize \textbf{93.8\%}} & {\scriptsize \textbf{98.1\%}} & {\scriptsize \textbf{98.8\%}}  \\ 

        \bottomrule
    \end{tabular}}
    \caption{
    Effect of \#slots $s$ and top-$k$ on LLaVA-1.5. Different combinations of $s$ - $k$ determine the retained \#tokens.
    }
    \label{tab:ablation_topk}
\end{table}

%% file: tables/ablation_layer.tex
\begin{table}
  \centering \small
  \begin{tabular}{lccc}
    \toprule
    $s$ & Layer 9 & Layer 10 & Layer -2 \\
    \midrule
    192 & 99.9\% & 99.9\% & 98.1\% \\
    128 & 97.5\% & 97.2\% & 95.8\% \\
    64 & 94.5\% & 93.6\% & 93.5\%\\
    \bottomrule
  \end{tabular}
  \caption{
  Performance of OC-VTP at different layers.
  Relative performance on LLaVA-1.5 averaged over the seven benchmarks (GQA, MMB, MME, POPE, VQA$^{\text{Text}}$, MMMU, and SEED).
  }
  \label{tab:ablation_layer}
\end{table}

%% file: tables/ablation_loss.tex
\begin{table}
  \centering \small
  \begin{tabular}{lccc}
    \toprule
    $s$ & MSE & AW-MSE \\
    \midrule
    192 & 99.3\% & 99.9\%  \\
    128 & 96.8\% & 97.5\%  \\
    64 & 92.4\% & 94.5\% \\
    \bottomrule
  \end{tabular}
  \caption{
  Effect of OC-pruner training losses.
  Relative performance on LLaVA-1.5 averaged over the seven benchmarks (GQA, MMB, MME, POPE, VQA$^{\text{Text}}$, MMMU, and SEED). AW-MSE is always better than MSE at different token budgets.
  }
  \label{tab:ablation_loss}
\end{table}

%% file: sec/5_conclusion.tex
\section{Conclusion}
\label{sect:conclusion}
We introduce OC-VTP, an Object-Centric Learning-based Vision Token Pruning method for VLMs. We prune vision tokens by utilizing Object-Centric Learning especially the Slot Attention module to select the most representative vision tokens, with guarantee for the first time.
Evaluated on LLaVA-1.5, LLaVA-NeXT, and Qwen2.5-VL, OC-VTP delivers great FLOPs savings and competitive latency compared to existing strong baselines, while preserving accuracy. In future work, we will extend OC-VTP to video and other tasks, and explore pruning in the decoder as well as text-token-assisted vision tokens pruning modules.

%% file: main.bib
@String(CVPR= {IEEE Conf. Comput. Vis. Pattern Recog.})

@String(AAAI = {AAAI})

@String(CVPR  = {CVPR})

@inproceedings{llava,
 author = {Liu, Haotian and Li, Chunyuan and Wu, Qingyang and Lee, Yong Jae},
 booktitle = {Advances in Neural Information Processing Systems},
 editor = {A. Oh and T. Naumann and A. Globerson and K. Saenko and M. Hardt and S. Levine},
 pages = {34892--34916},
 publisher = {Curran Associates, Inc.},
 title = {Visual Instruction Tuning},
 url = {https://proceedings.neurips.cc/paper_files/paper/2023/file/6dcf277ea32ce3288914faf369fe6de0-Paper-Conference.pdf},
 volume = {36},
 year = {2023}
}

@misc{llavanext,
  title={Llavanext: Improved reasoning, ocr, and world knowledge},
  author={Liu, Haotian and Li, Chunyuan and Li, Yuheng and Li, Bo and Zhang, Yuanhan and Shen, Sheng and Lee, Yong Jae},
  year={2024}
}

@misc{qwen2.5-vl,
      title={Qwen2.5-VL Technical Report}, 
      author={Shuai Bai and Keqin Chen and Xuejing Liu and Jialin Wang and Wenbin Ge and Sibo Song and Kai Dang and Peng Wang and Shijie Wang and Jun Tang and Humen Zhong and Yuanzhi Zhu and Mingkun Yang and Zhaohai Li and Jianqiang Wan and Pengfei Wang and Wei Ding and Zheren Fu and Yiheng Xu and Jiabo Ye and Xi Zhang and Tianbao Xie and Zesen Cheng and Hang Zhang and Zhibo Yang and Haiyang Xu and Junyang Lin},
      year={2025},
      eprint={2502.13923},
      archivePrefix={arXiv},
      primaryClass={cs.CV},
      url={https://arxiv.org/abs/2502.13923}, 
}

@article{mini-gemini,
  title={Mini-gemini: Mining the potential of multi-modality vision language models},
  author={Li, Yanwei and Zhang, Yuechen and Wang, Chengyao and Zhong, Zhisheng and Chen, Yixin and Chu, Ruihang and Liu, Shaoteng and Jia, Jiaya},
  journal={IEEE Transactions on Pattern Analysis and Machine Intelligence},
  year={2025},
  publisher={IEEE}
}

@inproceedings{mini-gpt,
  title={MiniGPT-4: Enhancing Vision-Language Understanding with Advanced Large Language Models},
  author={Zhu, Deyao and Chen, Jun and Shen, Xiaoqian and Li, Xiang and Elhoseiny, Mohamed},
  booktitle={The Twelfth International Conference on Learning Representations},
  year={2024}
}

@article{hiprune,
  author={Liu, Jizhihui and Du, Feiyi and Zhu, Guangdao and Lian, Niu and Li, Jun and Chen, Bin},
  title={HiPrune: Training-Free Visual Token Pruning via Hierarchical Attention in Vision-Language Models},
  journal={arXiv preprint arXiv:2508.00553},
  year={2025}
}

@inproceedings{ToMe,
  title={Token Merging: Your ViT But Faster},
  author={Bolya, Daniel and Fu, Cheng-Yang and Dai, Xiaoliang and Zhang, Peizhao and Feichtenhofer, Christoph and Hoffman, Judy},
  booktitle={The Eleventh International Conference on Learning Representations},
  year={2023}
}

@inproceedings{FastV,
  title={An image is worth 1/2 tokens after layer 2: Plug-and-play inference acceleration for large vision-language models},
  author={Chen, Liang and Zhao, Haozhe and Liu, Tianyu and Bai, Shuai and Lin, Junyang and Zhou, Chang and Chang, Baobao},
  booktitle={European Conference on Computer Vision},
  pages={19--35},
  year={2024},
  organization={Springer}
}

@inproceedings{TRIM,
  title={Less is More: A Simple yet Effective Token Reduction Method for Efficient Multi-modal LLMs},
  author={Song, Dingjie and Wang, Wenjun and Chen, Shunian and Wang, Xidong and Guan, Michael X and Wang, Benyou},
  booktitle={Proceedings of the 31st International Conference on Computational Linguistics},
  pages={7614--7623},
  year={2025}
}

@inproceedings{VisionZip,
  title={Visionzip: Longer is better but not necessary in vision language models},
  author={Yang, Senqiao and Chen, Yukang and Tian, Zhuotao and Wang, Chengyao and Li, Jingyao and Yu, Bei and Jia, Jiaya},
  booktitle={Proceedings of the Computer Vision and Pattern Recognition Conference},
  pages={19792--19802},
  year={2025}
}

@inproceedings{SparseVLM,
  title={SparseVLM: Visual Token Sparsification for Efficient Vision-Language Model Inference},
  author={Zhang, Yuan and Fan, Chun-Kai and Ma, Junpeng and Zheng, Wenzhao and Huang, Tao and Cheng, Kuan and Gudovskiy, Denis A and Okuno, Tomoyuki and Nakata, Yohei and Keutzer, Kurt and others},
  booktitle={International Conference on Machine Learning},
  pages={74840--74857},
  year={2025},
  organization={PMLR}
}

@InProceedings{PDrop,
    author    = {Xing, Long and Huang, Qidong and Dong, Xiaoyi and Lu, Jiajie and Zhang, Pan and Zang, Yuhang and Cao, Yuhang and He, Conghui and Wang, Jiaqi and Wu, Feng and Lin, Dahua},
    title     = {Conical Visual Concentration for Efficient Large Vision-Language Models},
    booktitle = {Proceedings of the Computer Vision and Pattern Recognition Conference (CVPR)},
    month     = {June},
    year      = {2025},
    pages     = {14593-14603}
}

@inproceedings{CLIP,
  title={Learning transferable visual models from natural language supervision},
  author={Radford, Alec and Kim, Jong Wook and Hallacy, Chris and Ramesh, Aditya and Goh, Gabriel and Agarwal, Sandhini and Sastry, Girish and Askell, Amanda and Mishkin, Pamela and Clark, Jack and others},
  booktitle={International conference on machine learning},
  pages={8748--8763},
  year={2021},
  organization={PmLR}
}

@article{ROPE,
  title={Roformer: Enhanced transformer with rotary position embedding},
  author={Su, Jianlin and Ahmed, Murtadha and Lu, Yu and Pan, Shengfeng and Bo, Wen and Liu, Yunfeng},
  journal={Neurocomputing},
  volume={568},
  pages={127063},
  year={2024},
  publisher={Elsevier}
}

@article{Qwen2.0,
  title={Qwen2-vl: Enhancing vision-language model's perception of the world at any resolution},
  author={Wang, Peng and Bai, Shuai and Tan, Sinan and Wang, Shijie and Fan, Zhihao and Bai, Jinze and Chen, Keqin and Liu, Xuejing and Wang, Jialin and Ge, Wenbin and others},
  journal={arXiv preprint arXiv:2409.12191},
  year={2024}
}

@inproceedings{mmb,
  title={Mmbench: Is your multi-modal model an all-around player?},
  author={Liu, Yuan and Duan, Haodong and Zhang, Yuanhan and Li, Bo and Zhang, Songyang and Zhao, Wangbo and Yuan, Yike and Wang, Jiaqi and He, Conghui and Liu, Ziwei and others},
  booktitle={European conference on computer vision},
  pages={216--233},
  year={2024},
  organization={Springer}
}

@inproceedings{gqa,
  title={Gqa: A new dataset for real-world visual reasoning and compositional question answering},
  author={Hudson, Drew A and Manning, Christopher D},
  booktitle={Proceedings of the IEEE/CVF conference on computer vision and pattern recognition},
  pages={6700--6709},
  year={2019}
}

@inproceedings{pope,
  title={Evaluating Object Hallucination in Large Vision-Language Models},
  author={Li, Yifan and Du, Yifan and Zhou, Kun and Wang, Jinpeng and Zhao, Xin and Wen, Ji-Rong},
  booktitle={The 2023 Conference on Empirical Methods in Natural Language Processing},
  year={2023}
}

@inproceedings{textVQA,
  title={Towards vqa models that can read},
  author={Singh, Amanpreet and Natarajan, Vivek and Shah, Meet and Jiang, Yu and Chen, Xinlei and Batra, Dhruv and Parikh, Devi and Rohrbach, Marcus},
  booktitle={Proceedings of the IEEE/CVF conference on computer vision and pattern recognition},
  pages={8317--8326},
  year={2019}
}

@article{sqa,
  title={Learn to explain: Multimodal reasoning via thought chains for science question answering},
  author={Lu, Pan and Mishra, Swaroop and Xia, Tanglin and Qiu, Liang and Chang, Kai-Wei and Zhu, Song-Chun and Tafjord, Oyvind and Clark, Peter and Kalyan, Ashwin},
  journal={Advances in Neural Information Processing Systems},
  volume={35},
  pages={2507--2521},
  year={2022}
}

@article{slotAttention,
  title={Object-centric learning with slot attention},
  author={Locatello, Francesco and Weissenborn, Dirk and Unterthiner, Thomas and Mahendran, Aravindh and Heigold, Georg and Uszkoreit, Jakob and Dosovitskiy, Alexey and Kipf, Thomas},
  journal={Advances in neural information processing systems},
  volume={33},
  pages={11525--11538},
  year={2020}
}

@inproceedings{DINOSAUR,
  title={Bridging the Gap to Real-World Object-Centric Learning},
  author={Seitzer, Maximilian and Horn, Max and Zadaianchuk, Andrii and Zietlow, Dominik and Xiao, Tianjun and Simon-Gabriel, Carl-Johann and He, Tong and Zhang, Zheng and Sch{\"o}lkopf, Bernhard and Brox, Thomas and others},
  booktitle={The Eleventh International Conference on Learning Representations},
  year={2023}
}

@inproceedings{COCO,
  title={Microsoft coco: Common objects in context},
  author={Lin, Tsung-Yi and Maire, Michael and Belongie, Serge and Hays, James and Perona, Pietro and Ramanan, Deva and Doll{\'a}r, Piotr and Zitnick, C Lawrence},
  booktitle={European conference on computer vision},
  pages={740--755},
  year={2014},
  organization={Springer}
}

@inproceedings{lmms,
  title={Lmms-eval: Reality check on the evaluation of large multimodal models},
  author={Zhang, Kaichen and Li, Bo and Zhang, Peiyuan and Pu, Fanyi and Cahyono, Joshua Adrian and Hu, Kairui and Liu, Shuai and Zhang, Yuanhan and Yang, Jingkang and Li, Chunyuan and others},
  booktitle={Findings of the Association for Computational Linguistics: NAACL 2025},
  pages={881--916},
  year={2025}
}

@inproceedings{mme,
  title={MME: A Comprehensive Evaluation Benchmark for Multimodal Large Language Models},
  author={Fu, Chaoyou and Chen, Peixian and Shen, Yunhang and Qin, Yulei and Zhang, Mengdan and Lin, Xu and Yang, Jinrui and Zheng, Xiawu and Li, Ke and Sun, Xing and others},
  booktitle={The Thirty-ninth Annual Conference on Neural Information Processing Systems Datasets and Benchmarks Track},
  year={2025}
}

@inproceedings{vqav2,
  title={Making the v in vqa matter: Elevating the role of image understanding in visual question answering},
  author={Goyal, Yash and Khot, Tejas and Summers-Stay, Douglas and Batra, Dhruv and Parikh, Devi},
  booktitle={Proceedings of the IEEE conference on computer vision and pattern recognition},
  pages={6904--6913},
  year={2017}
}

@inproceedings{vizwiz,
  title={Vizwiz grand challenge: Answering visual questions from blind people},
  author={Gurari, Danna and Li, Qing and Stangl, Abigale J and Guo, Anhong and Lin, Chi and Grauman, Kristen and Luo, Jiebo and Bigham, Jeffrey P},
  booktitle={Proceedings of the IEEE conference on computer vision and pattern recognition},
  pages={3608--3617},
  year={2018}
}

@inproceedings{mmmu,
  title={Mmmu: A massive multi-discipline multimodal understanding and reasoning benchmark for expert agi},
  author={Yue, Xiang and Ni, Yuansheng and Zhang, Kai and Zheng, Tianyu and Liu, Ruoqi and Zhang, Ge and Stevens, Samuel and Jiang, Dongfu and Ren, Weiming and Sun, Yuxuan and others},
  booktitle={Proceedings of the IEEE/CVF Conference on Computer Vision and Pattern Recognition},
  pages={9556--9567},
  year={2024}
}

@inproceedings{seedBench,
  title={Seed-bench: Benchmarking multimodal large language models},
  author={Li, Bohao and Ge, Yuying and Ge, Yixiao and Wang, Guangzhi and Wang, Rui and Zhang, Ruimao and Shan, Ying},
  booktitle={Proceedings of the IEEE/CVF Conference on Computer Vision and Pattern Recognition},
  pages={13299--13308},
  year={2024}
}

@article{wu2023slotdiffuz,
  title={{SlotDiffusion: Object-Centric Generative Modeling with Diffusion Models}},
  author={Wu, Ziyi and Hu, Jingyu and Lu, Wuyue and Gilitschenski, Igor and Garg, Animesh},
  journal={Advances in Neural Information Processing Systems},
  volume={36},
  pages={50932--50958},
  year={2023}
}

@inproceedings{kakogeorgiou2024spot,
  title={{Spot: Self-Training with Patch-Order Permutation for Object-Centric Learning with Autoregressive Transformers}},
  author={Kakogeorgiou, Ioannis and Gidaris, Spyros and Karantzalos, Konstantinos and Komodakis, Nikos},
  booktitle={Proceedings of the IEEE/CVF Conference on Computer Vision and Pattern Recognition},
  pages={22776--22786},
  year={2024}
}

@inproceedings{zhao2025vvo,
  title={Vector-quantized vision foundation models for object-centric learning},
  author={Zhao, Rongzhen and Wang, Vivienne Huiling and Kannala, Juho and Pajarinen, Joni},
  booktitle={Proceedings of the 33rd ACM International Conference on Multimedia},
  pages={5422--5430},
  year={2025}
}

@inproceedings{zhao2025dias,
  title={Slot attention with re-initialization and self-distillation},
  author={Zhao, Rongzhen and Zhao, Yi and Kannala, Juho and Pajarinen, Joni},
  booktitle={Proceedings of the 33rd ACM International Conference on Multimedia},
  pages={4185--4192},
  year={2025}
}

@inproceedings{zhao2025randsfq,
  title={Predicting video slot attention queries from random slot-feature pairs},
  author={Zhao, Rongzhen and Li, Jian and Kannala, Juho and Pajarinen, Joni},
  booktitle={Proceedings of the AAAI Conference on Artificial Intelligence},
  volume={40},
  number={16},
  pages={13208--13216},
  year={2026}
}

@article{zhao2025smoothsa,
  title={Smoothing Slot Attention Iterations and Recurrences},
  author={Zhao, Rongzhen and Yang, Wenyan and Kannala, Juho and Pajarinen, Joni},
  journal={arXiv preprint arXiv:2508.05417},
  year={2025}
}

@article{zadaianchuk2024videosaur,
  title={Object-centric learning for real-world videos by predicting temporal feature similarities},
  author={Zadaianchuk, Andrii and Seitzer, Maximilian and Martius, Georg},
  journal={Advances in neural information processing systems},
  volume={36},
  pages={61514--61545},
  year={2023}
}

@inproceedings{manasyan2025slotcontrast,
  title={Temporally consistent object-centric learning by contrasting slots},
  author={Manasyan, Anna and Seitzer, Maximilian and Radovic, Filip and Martius, Georg and Zadaianchuk, Andrii},
  booktitle={Proceedings of the Computer Vision and Pattern Recognition Conference},
  pages={5401--5411},
  year={2025}
}

@inproceedings{jia2023boqsa,
  title={Improving Object-centric Learning with Query Optimization},
  author={Jia, Baoxiong and Liu, Yu and Huang, Siyuan},
  booktitle={The Eleventh International Conference on Learning Representations},
  year={2023}
}

@inproceedings{biza2023isa,
  title={Invariant slot attention: object discovery with slot-centric reference frames},
  author={Biza, Ondrej and Van Steenkiste, Sjoerd and Sajjadi, Mehdi SM and Elsayed, Gamaleldin F and Mahendran, Aravindh and Kipf, Thomas},
  booktitle={Proceedings of the 40th International Conference on Machine Learning},
  pages={2507--2527},
  year={2023}
}
